\documentclass{article}

\usepackage[preprint]{neurips_2026}

\usepackage[utf8]{inputenc} 
\usepackage[T1]{fontenc}    
\usepackage[hidelinks]{hyperref}       
\usepackage{url}            
\usepackage{booktabs}       
\usepackage{amsfonts}       
\usepackage{nicefrac}       
\usepackage{microtype}      
\usepackage{xcolor}         
\usepackage{graphicx}
\usepackage{subcaption}

\usepackage{amsmath}
\usepackage{amssymb}
\usepackage{mathtools}
\usepackage{amsthm}

\usepackage{algorithm}
\usepackage{algorithmic}

\usepackage{wrapfig}
\usepackage[table]{xcolor}
\definecolor{best}{RGB}{255, 204, 214}    
\definecolor{second}{RGB}{255, 235, 238}  
\usepackage[most]{tcolorbox}
\usepackage{enumitem}
\usepackage{multirow}

\setcitestyle{numbers,square,comma,sort&compress}

\title{Risk Horizons: Structured Hypothesis Spaces for Longitudinal Clinical Prediction}

\author{%
  Zhan Qu \\
  TU Dresden and ScaDS.AI \\
  Dresden, Germany \\
  \texttt{zhan.qu@tu-dresden.de} \\
  \And
  Michael Färber \\
  TU Dresden and ScaDS.AI \\
  Dresden, Germany \\
  \texttt{michael.faerber@tu-dresden.de} \\
}

\begin{document}

\maketitle

\begin{abstract}
  Predicting future clinical events from longitudinal electronic health records (EHRs) requires selecting plausible outcomes from a large and structured event space under sparse observations. While clinical coding systems provide hierarchical organization of events, cross-modal and temporal relationships are not explicitly specified and must instead be inferred from data, making prediction difficult for weakly observed longitudinal transitions. We introduce \textbf{Risk Horizons}, a geometry-aware framework for constructing patient-specific candidate spaces for multi-modal next-visit prediction. Risk Horizons combines deterministic coding hierarchies with data-driven lagged cross-modal associations, embeds the resulting clinical graph in hyperbolic space, and retrieves candidate futures using directional risk cones. This reframes longitudinal prediction as ranking within a compact, clinically coherent hypothesis space rather than scoring an unconstrained vocabulary. Experiments on MIMIC-IV and eICU demonstrate competitive next-visit prediction performance, with consistently improved hierarchy consistency across diagnoses, procedures, and medications. Further analysis suggests that hyperbolic structured candidate retrieval is the primary driver of performance, while LLMs are effective as constrained inference-time rerankers operating over clinically grounded candidate sets.
\end{abstract}

\section{Introduction}
\label{sec:intro}
Longitudinal electronic health records (EHRs) are central to clinical decision support, enabling models of disease progression, treatment response, and adverse events over time \citep{rajkomar2018scalable,harutyunyan2019multitask,choi2016doctor}. A standard formulation is next-visit prediction, where the goal is to forecast a patient’s future diagnoses, procedures, and medications from prior visits. Existing approaches commonly represent patient histories as temporal sequences and apply recurrent or transformer-based architectures to predict future clinical events \citep{choi2016retain,li2020behrt}. More recently, large language models (LLMs) have been explored for clinical reasoning due to their strong semantic representations and few-shot generalization capabilities \citep{singhal2023large,nori2023capabilities}. However, longitudinal EHR prediction remains difficult because the space of plausible future events is large, sparse, and strongly structured.

A key difficulty is that EHR trajectories are sparse, irregularly sampled, and multi-modal \citep{lipton2015learning,che2018recurrent}. Within a single admission, information is recorded as multiple unordered sets of entities drawn from different coding systems \citep{johnson2023mimic}, such as diagnoses in ICD and medications in NDC/ATC, rather than as a coherent sequence of clinically salient states. This fragmentation makes it unclear which elements within a visit best summarize the patient’s condition, and it leaves many clinically meaningful relationships implicit, such as which diagnoses motivate which treatments.

Clinical knowledge resources provide partial structure for this problem, but not the full hypothesis space required for longitudinal prediction. External biomedical knowledge bases such as UMLS \citep{bodenreider2004unified} provide broad relations and hierarchies, but can be difficult to operationalize at scale due to heterogeneous vocabularies and weakly specified relations. In contrast, coding systems such as ICD-10 \citep{world2004international} and the Anatomical Therapeutic Chemical classification \citep{who_atc} provide reliable hierarchical organization of clinical events. However, these hierarchies do not explicitly encode cross-modal or temporal relationships among diagnoses, procedures, and medications. Such dependencies must therefore be inferred from longitudinal data, where they are often sparse or weakly observed.

\begin{figure}[!t] 
    \centering 
    \includegraphics[width=1\linewidth]{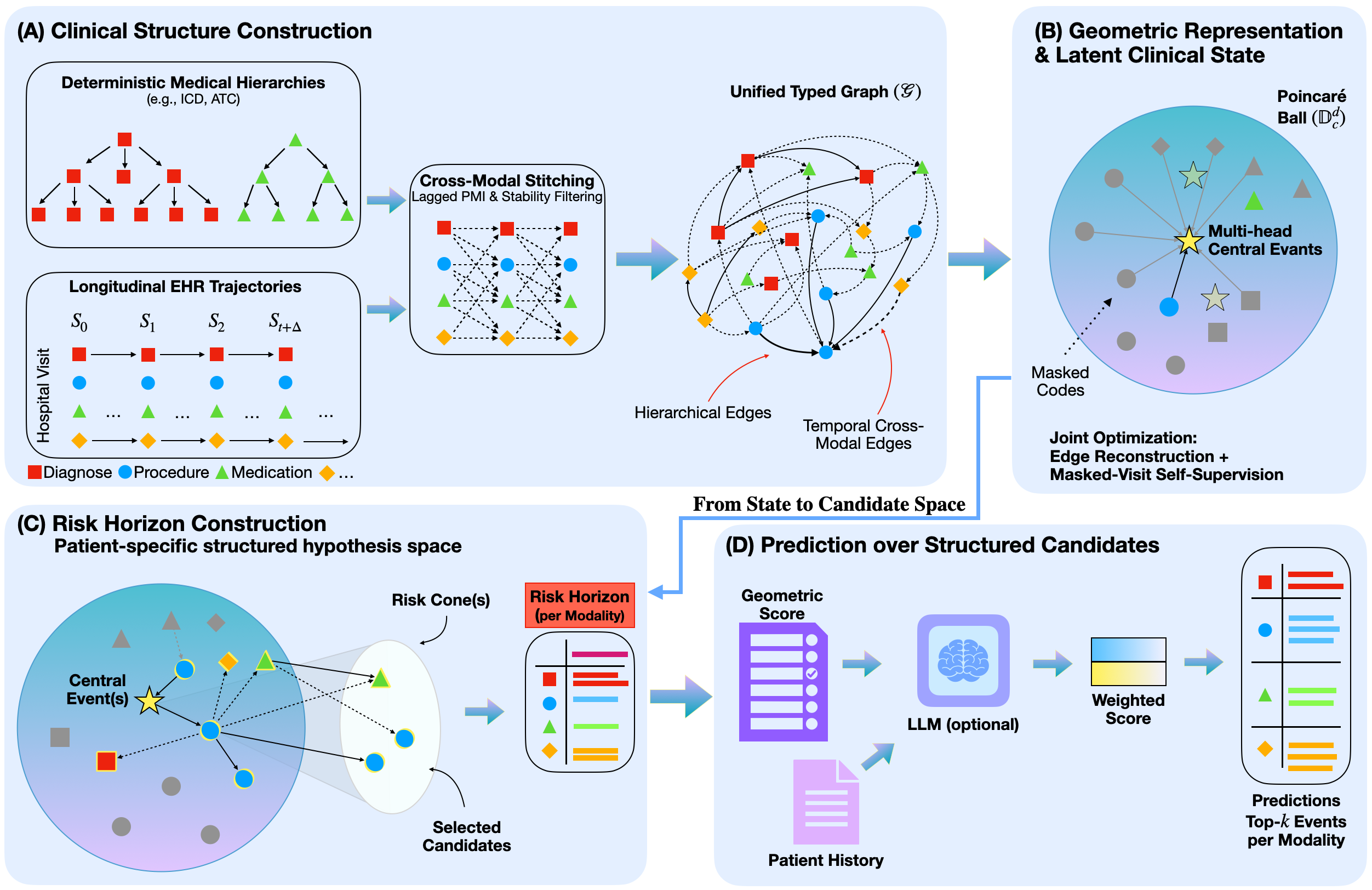} 
    \vspace{-0.3cm}
    \caption{\textbf{Overview of Risk Horizons.}
    (A) Clinical structure is constructed by combining coding hierarchies with data-driven cross-modal associations.
    (B) Concepts are embedded in hyperbolic space, and visits are summarized as multi-head Central Events.
    (C) Structure-aware retrieval constructs a patient-specific \emph{Risk Horizon}, a set of plausible future events.
    (D) Prediction ranks candidates within the Risk Horizon using geometric scores, optionally refined by an LLM.
    }
    \label{fig:overview} 
    \vspace{-0.6cm}
\end{figure}

This motivates a different view of next-visit prediction. Rather than treating the task only as reasoning over patient history, we view it as constructing a patient-specific hypothesis space of plausible futures. In clinical practice, decision-making does not proceed by scoring over all possible diagnoses or treatments. Instead, clinicians operate within a context-dependent space of plausible alternatives constrained by prior knowledge, hierarchical relationships, and disease progression. This suggests that models should first recover clinically coherent candidate futures and only then rank predictions within this space. This is particularly important for sparse and weakly supported transitions, where full-vocabulary scoring or unstructured retrieval often yields diffuse or clinically inconsistent predictions.

We introduce \textbf{Risk Horizons}, a geometry-aware framework for constructing patient-specific structured candidate spaces for longitudinal EHR prediction. As shown in Figure~\ref{fig:overview}, Risk Horizons builds a unified clinical graph by combining deterministic coding-system hierarchies with data-driven lagged cross-modal associations. The graph is embedded in hyperbolic space, whose geometry is well suited to hierarchical abstraction and asymmetric relationships \citep{nickel2017poincare,ganea2018hyperbolic}. Patient visits are summarized as probabilistic \emph{Central Events}, which denoise sparse multi-modal observations into latent clinical states. At inference time, directional geometric \emph{risk cones} retrieve compact sets of clinically plausible future events aligned with hierarchy and inferred temporal progression.

This formulation turns next-visit prediction into a two-stage procedure: first construct a structured Risk Horizon, then rank candidates within it. LLMs are used only as optional inference-time rerankers over retrieved candidates, not as unconstrained generators. This design preserves the semantic strengths of LLMs while restricting their output space to clinically plausible candidates, reducing the likelihood of clinically implausible generations \citep{ji2023survey,zhou2025evaluating,qu2025medieval}. Our experiments suggest that geometry-aware candidate construction contributes substantially to performance, while LLM reranking provides additional gains when constrained to the Risk Horizon.

\textbf{Our contributions are:} \textbf{(1)} we reframe longitudinal EHR prediction as structured hypothesis space construction rather than only sequence modeling or unconstrained reasoning; \textbf{(2)} we introduce Risk Horizons, a geometry-aware framework that combines deterministic coding hierarchies with data-driven lagged cross-modal associations; \textbf{(3)} we propose directional retrieval via geometric risk cones to construct compact, patient-specific candidate spaces; and \textbf{(4)} we provide extensive evaluation on MIMIC-IV and eICU, demonstrating strong next-visit prediction performance and improved hierarchy consistency under partially observed longitudinal structure. Our code is publicly available at: \href{https://anonymous.4open.science/r/Risk-Horizons-6557/}{https://anonymous.4open.science/r/Risk-Horizons-6557/}.

\section{Related Work}
\label{sec:related_work}
\textbf{Deep Sequential Models for Longitudinal EHRs.}
Longitudinal EHRs are commonly modeled as temporal sequences. Early recurrent models demonstrated the feasibility of next-visit prediction from diagnosis codes \citep{choi2016doctor, pham2016deepcare}, while RETAIN introduced reverse-time attention for interpretability \citep{choi2016retain} and later work incorporated uncertainty-aware and survival objectives \citep{alaa2019attentive}. Transformer-based models treat medical codes as tokens to capture long-range dependencies \citep{li2020behrt, rasmy2021med}, with recent extensions exploring long-context, encoder--decoder, and foundation-model pretraining for irregular clinical trajectories \citep{wornow2024context, yang2023transformehr, steinberg2023motor}. Autoregressive models such as HALO further model high-dimensional EHR sequences \citep{theodorou2023synthesize}. While effective at modeling temporal patterns, these approaches largely treat prediction as sequence modeling over flattened event sets, without explicitly constructing structured candidate spaces for plausible future events.

\textbf{Ontology-Aware and Graph-Based Clinical Representation Learning.}
Ontology-aware methods incorporate medical hierarchies such as ICD and SNOMED-CT. GRAM propagates information from ancestor concepts via attention \citep{choi2017gram}, while KAME integrates external knowledge graphs into memory networks \citep{ma2018kame}. Graph-based approaches construct patient--code or code--code graphs and apply GNNs for clinical representation learning \citep{shang2019pre, mao2022medgcn}, with recent work extending this direction to temporal heterogeneous graphs using time-aware message passing or graph transformers \citep{xu2022time, chen2024predictive}. These methods embed relational and ontological structure into learned representations, but primarily use this structure for representation learning or downstream prediction rather than explicitly constructing constrained candidate spaces for future events.

\textbf{Hyperbolic Geometry for Medical Representation Learning.}
Hyperbolic geometry provides low-distortion embeddings for hierarchical and tree-structured data due to exponential volume growth \citep{sarkar2011low, nickel2017poincare}. Hyperbolic entailment cones and hyperbolic GNNs extend these ideas to non-Euclidean neural models \citep{ganea2018hyperbolic, chami2019hyperbolic, liu2019hyperbolic}. In medicine, hyperbolic embeddings have been used to capture ICD hierarchies and improve mortality, readmission, diagnosis prediction, and ICD coding tasks \citep{beaulieu2019learning, lu2019learning, lu2021self, naseem2024graph, cao2020hypercore, wu2024hyperbolic}. Prior work primarily uses hyperbolic geometry for representation learning or classification, and does not exploit its geometric structure to construct structured hypothesis spaces for downstream prediction.

\textbf{RAG and LLM-Based Clinical Reasoning.} LLMs show strong performance on medical reasoning benchmarks \citep{singhal2023large, nori2023capabilities}, but hallucination remains a concern in clinical settings \citep{ji2023survey}. Retrieval-augmented generation grounds predictions in retrieved evidence \citep{lewis2020retrieval, guu2020retrieval}, and clinical RAG studies show that retrieval strategy and corpus choice affect performance \citep{xiong2024benchmarking}. Beyond text retrieval, knowledge graph--augmented prompting retrieves entities, relations, or subgraphs \citep{pan2024unifying, zeng2025kosel}, and graph-based retrieval has been integrated into medical RAG pipelines \citep{wu2025medical}. Recent systems for longitudinal EHR reasoning, such as TRACE, maintain structured patient states over time \citep{qu2026trace}. Most retrieval mechanisms, however, rely on lexical similarity, nearest-neighbor retrieval, or linearized structured evidence, limiting their ability to model hierarchical abstraction and directional clinical progression.

\section{Methodology}
\label{sec:methodology}

We present \textbf{Risk Horizons}, a geometry-aware framework for constructing structured candidate spaces for longitudinal EHR prediction. The central challenge in next-visit prediction is that the space of possible future clinical events is only partially structured: while coding systems define hierarchical relationships, many clinically meaningful temporal and cross-modal dependencies remain implicit in longitudinal data. This yields a large and weakly constrained prediction space that is difficult to reason over directly. Risk Horizons addresses this challenge by constructing patient-specific candidate spaces that recover missing structure and constrain prediction to clinically plausible futures before final ranking. Figure~\ref{fig:overview} summarizes the framework, and Algorithm~\ref{alg:RiskHorizons} in Appendix~\ref{app:algorithm} provides the end-to-end procedure.

\subsection{Problem Setup and Notation}
\label{subsec:problem_setup}

A patient trajectory is a sequence of visits indexed by time $t\in\{1,\dots,T\}$. Each visit is a set-valued, multi-modal observation:
\begin{equation}
S_t=\left\{S_t^{(\mathrm{dx})},\,S_t^{(\mathrm{proc})},\,S_t^{(\mathrm{med})}\right\},
\qquad
S_t^{(m)}\subset \mathcal{V}^{(m)}.
\label{eq:visit_def}
\end{equation}

Given history $S_{1:t}$, the goal is to predict next-visit events $S_{t+1}^{(m)}$ for each modality $m$. We denote the unified vocabulary by
\begin{equation}
\mathcal{V}=\bigcup_m \mathcal{V}^{(m)}\cup \mathcal{V}^{(\mathrm{anc})},
\label{eq:vocab_def}
\end{equation}
where $\mathcal{V}^{(\mathrm{anc})}$ contains deterministic ancestor nodes induced by coding hierarchies. We formulate prediction as ranking events within a patient-specific structured candidate space $\mathcal{R}_T\subset\mathcal{V}$, referred to as the \emph{Risk Horizon}. The Risk Horizon is intended to capture clinically plausible future events before final modality-wise ranking.

\subsection{Risk Horizons: Structured Candidate Space Construction}
\label{subsec:risk_horizons}

\paragraph{Unified Clinical Graph.}
We construct a directed typed graph $\mathcal{G}=(\mathcal{V},\mathcal{E})$ with two edge families: (i) deterministic hierarchical edges and (ii) data-driven temporal cross-modal edges. Edges are partitioned by relation types $r\in\mathcal{R}$:
\begin{equation}
\mathcal{E}=\mathcal{E}_{\mathrm{hier}}\cup \mathcal{E}_{\mathrm{cross}},
\qquad
\mathcal{E}=\bigcup_{r\in\mathcal{R}} \mathcal{E}_r.
\label{eq:edge_partition}
\end{equation}

\paragraph{Hierarchical edges.}
Within each modality, we build deterministic parent--child edges from coding structure (e.g., ICD prefixes, ATC levels). We direct edges from abstract parents to specific children:
\begin{equation}
\mathcal{E}_{\mathrm{hier}}=\left\{(p\rightarrow v)\,:\, p=\mathrm{parent}(v),\; v\in \mathcal{V}^{(m)}\right\}.
\label{eq:hier_edges}
\end{equation}
This directionality ensures that specialization corresponds to moving ``downstream'' in the hierarchy.

\paragraph{Temporal Cross-Modal Stitching via Lagged PMI.}
To connect modality trees and encode longitudinal dependencies not explicitly specified by coding hierarchies, we define lagged co-occurrence for $a\in\mathcal{V}^{(m)}$ and $b\in\mathcal{V}^{(m')}$, $m\neq m'$:
\begin{equation}
P_{\Delta}(a,b)=\Pr\!\left(a\in S_t^{(m)},\; b\in S_{t+\Delta}^{(m')}\right),
\qquad \Delta\in\{0,\dots,\Delta_{\max}\}.
\label{eq:lag_cooc}
\end{equation}

We compute lagged PMI (associational, not causal):
\begin{equation}
\mathrm{PMI}_{\Delta}(a\rightarrow b)=\log \frac{P_{\Delta}(a,b)}{P(a)P(b)}.
\label{eq:lag_pmi}
\end{equation}
We retain a directed cross-modal edge $a\rightarrow b$ if
\begin{equation}
\mathrm{PMI}_{\Delta}(a\rightarrow b)>\tau_{m,m',\Delta}
\quad \wedge \quad
\mathrm{cnt}_{\Delta}(a,b)\ge \kappa,
\label{eq:pmi_filter}
\end{equation}
where $\kappa$ is a minimum support threshold. Each retained edge is assigned a type
\begin{equation}
r=\mathrm{type}(a\rightarrow b)=(m\rightarrow m',\Delta),
\label{eq:edge_type}
\end{equation}
yielding $\mathcal{E}_{\mathrm{cross}}=\bigcup_r \mathcal{E}_r$.

\paragraph{Stability Filtering.}
To reduce dataset-specific artifacts, we apply bootstrap stability filtering to PMI-derived edges. Let $\mathcal{E}^{(b)}_{\mathrm{cross}}$ denote the cross-modal edge set recovered on bootstrap resample $b\in\{1,\dots,B\}$. We keep edge $e$ iff
\begin{equation}
\mathrm{stab}(e)=\frac{1}{B}\sum_{b=1}^{B}\mathbb{I}\!\left[e\in \mathcal{E}^{(b)}_{\mathrm{cross}}\right]\ge q,
\label{eq:stability}
\end{equation}
where $q\in(0,1]$ controls the stability requirement.

\paragraph{Hyperbolic Representation.}
We embed nodes into the $d$-dimensional Poincar\'e ball of curvature $-c$:
\begin{equation}
\mathbb{D}_c^d=\left\{x\in\mathbb{R}^d:\|x\|<1/\sqrt{c}\right\}, 
\qquad z_v\in \mathbb{D}_c^d\;\;\forall v\in\mathcal{V}.
\label{eq:poincare_ball}
\end{equation}
For any two embeddings $x,y\in\mathbb{D}_c^d$, we use the hyperbolic distance $d_{\mathbb{D}}(x,y)$ (Appendix~\ref{app:geometry}). For an observed edge $(u\rightarrow v)$ of type $r$, we define a compatibility score
\begin{equation}
s_r(u,v)=-d_{\mathbb{D}}(z_u,z_v).
\label{eq:compatibility}
\end{equation}

Embeddings are trained via typed edge reconstruction with negative sampling; the full typed edge reconstruction objective and optimization details are provided in Appendix~\ref{app:edge} and optimized with Riemannian SGD/Adam with projection to enforce $\|z_v\|<1/\sqrt{c}$.

\paragraph{Latent Clinical State via Probabilistic Central Events.}
Let $\mathcal{C}_t=\bigcup_m S_t^{(m)}$ be the set of all codes observed at visit $t$. We compute a visit-level hyperbolic barycenter as a closed-form approximation of the Fr\'echet mean using log--exp maps at the origin:
\begin{equation}
\mu_t
=
\exp_0^c\!\left(
\frac{1}{\sum_{c_i\in \mathcal{C}_t} w_{t,i}}
\sum_{c_i\in \mathcal{C}_t} w_{t,i}\log_0^c(z_{c_i})
\right),
\label{eq:barycenter_main}
\end{equation}
where $w_{t,i}\ge 0$ are code weights (uniform by default).

Rather than snapping $\mu_t$ to a single nearest node, we define a soft assignment over candidate nodes:
\begin{equation}
p_t(v)=
\frac{\exp\!\left(-\beta\, d_{\mathbb{D}}(\mu_t,z_v)\right)}
{\sum_{v'\in\mathcal{V}_t}\exp\!\left(-\beta\, d_{\mathbb{D}}(\mu_t,z_{v'})\right)},
\label{eq:soft_assign_main}
\end{equation}
where $\mathcal{V}_t$ is a restricted candidate pool (e.g., same-modality nodes and ancestors) for efficiency. The Central Event is $\mathrm{CE}_t=(\mu_t,p_t(\cdot))$.

\paragraph{Multi-head Central Events.}
A single visit barycenter may collapse heterogeneous clinical signals, e.g., diagnoses, procedures, and medications, into one overly coarse state that obscures distinct clinical factors. We therefore represent each visit by $H$ head-specific Central Events. For each code $c_i\in\mathcal{C}_t$, let $\tilde z_i=\log_0^c(z_{c_i})$. Each head applies a learned tangent-space projection $A_h\in\mathbb{R}^{d\times d}$ and computes
\begin{equation}
\bar u_t^{(h)}
=
\frac{\sum_{c_i\in\mathcal{C}_t} w_{t,i} A_h \tilde z_i}
{\sum_{c_i\in\mathcal{C}_t} w_{t,i}},
\qquad
\mu_t^{(h)}=\exp_0^c(\bar u_t^{(h)}).
\end{equation}
Each head induces a soft Central Event distribution
\begin{equation}
p_t^{(h)}(v)
=
\frac{\exp(-\beta d_{\mathbb D}(\mu_t^{(h)},z_v))}
{\sum_{v'\in\mathcal{V}_t}\exp(-\beta d_{\mathbb D}(\mu_t^{(h)},z_{v'}))}.
\end{equation}
The head projections are trained through masked-visit reconstruction, allowing different heads to specialize in distinct clinical factors. When a single visit summary is required, we use the mixture barycenter
\begin{equation}
\mu_t=\exp_0^c\!\left(\sum_{h=1}^H \pi_h \log_0^c(\mu_t^{(h)})\right),
\qquad
\sum_{h=1}^H\pi_h=1 .
\end{equation}
This reduces to the single-head formulation when $H=1$.

\paragraph{Masked-Visit Reconstruction for Self-Supervised Denoising.}
To train Central Events to preserve predictable clinical signal while filtering noise, we randomly mask a subset of visit codes $\mathcal{M}_t\subset \mathcal{C}_t$ and compute $\mu_t$ from $\mathcal{C}_t\setminus \mathcal{M}_t$, where $\mu_t$ is the mixture barycenter of the multi-head Central Events, which is used as a unified representation for reconstruction. We then score a masked code $c\in\mathcal{M}_t$ by hyperbolic distance:
\begin{equation}
p(c\mid \mu_t)\propto \exp\!\left(-d_{\mathbb{D}}(\mu_t,z_c)/\tau\right).
\label{eq:mask_prob_main}
\end{equation}
Training minimizes the sampled-softmax / contrastive reconstruction loss
\begin{equation}
\mathcal{L}_{\mathrm{mask}}
=
-\sum_{c\in\mathcal{M}_t}\log p(c\mid \mu_t),
\label{eq:mask_loss_main}
\end{equation}
with an efficient negative-sampling normalization. The complete sampled-softmax formulation and negative sampling strategy are detailed in Appendix~\ref{app:mask}, Eq.~\eqref{eq:mask_loss}.

\paragraph{Structure-Aware Retrieval via Multi-Cone Risk Horizons.}
Given the most recent Central Event at time $T$, Risk Horizons performs \emph{structure-aware retrieval} to identify a compact set of clinically plausible next-visit events. We refer to this retrieved, modality-typed candidate set as the \emph{Risk Horizon}. Unlike standard nearest-neighbor retrieval in embedding space, our approach explicitly incorporates hierarchical structure, cross-modal temporal associations, and directional progression. The goal is to favor candidates representing plausible downstream specialization or progression rather than merely nearby semantic similarity.

For each head $h$, we select a representative concept
$v_T^{(h)}=\arg\max_v p_T^{(h)}(v)$ and construct a structured candidate pool
\begin{equation}
\mathcal{C}_{T,h}^{\mathrm{cand}}
=
\mathrm{Desc}(v_T^{(h)})
\cup
\mathrm{Assoc}(v_T^{(h)})
\cup
\mathrm{Pred}(v_T^{(h)}),
\end{equation}
where $\mathrm{Desc}$ denotes hierarchy descendants, $\mathrm{Assoc}$ denotes same-visit cross-modal neighbors, and $\mathrm{Pred}$ denotes lagged temporal neighbors.

To favor clinically plausible downstream progression, each head defines a tangent-space risk cone. Let $r(v_T^{(h)})$ be the highest-level ancestor of $v_T^{(h)}$. The downstream direction is
\begin{equation}
d_T^{(h)}
=
\log_0^c(z_{v_T^{(h)}})
-
\log_0^c(z_{r(v_T^{(h)})}).
\end{equation}
A candidate $u$ is retained if
\begin{equation}
\angle(u;v_T^{(h)}) \le \phi,
\qquad
\cos\angle(u;v_T^{(h)})
=
\frac{\langle \log_0^c(z_u), d_T^{(h)}\rangle}
{\|\log_0^c(z_u)\|\|d_T^{(h)}\|}.
\end{equation}

Each retained candidate receives a head-specific geometric score
\begin{equation}
s_{\mathrm{geo}}^{(h)}(u;T)
=
- d_{\mathbb D}(z_u,\mu_T^{(h)})
+ \eta \cdot \mathbb{I}\!\left[u\in\mathrm{Cone}^{(h)}(v_T^{(h)})\right],
\end{equation}
and the final score aggregates across heads:
\begin{equation}
s_{\mathrm{geo}}(u;T)
=
\sum_{h=1}^H \pi_h\, s_{\mathrm{geo}}^{(h)}(u;T).
\end{equation}

The typed Risk Horizon $\mathcal{R}_T$ is formed by selecting the top-$K$ candidates per modality.

\subsection{Prediction over Structured Candidate Spaces}
\label{subsec:prediction}

Given the Risk Horizon $\mathcal{R}_T$, prediction is performed by ranking candidates within each modality. By default, we use the geometric score $s_{\mathrm{geo}}(u;T)$ from the Risk Horizon construction. Optionally, retrieved candidates can be reranked using a constrained LLM-based scoring module. Let $\mathrm{Desc}(\cdot)$ map codes to short text descriptions. The LLM receives (i) a compressed history from Central Events and (ii) candidates grouped by modality. For each candidate $u\in\mathcal{R}_T^{(m)}$, we obtain an LLM score $s_{\mathrm{llm}}(u;T)$ and combine it with the geometric score:
\begin{equation}
s(u;T)=\lambda\, s_{\mathrm{llm}}(u;T) + (1-\lambda)\, s_{\mathrm{geo}}(u;T).
\label{eq:rerank_main}
\end{equation}
When no LLM reranker is used, we set $s(u;T)=s_{\mathrm{geo}}(u;T)$. The LLM is used strictly at inference time and is not allowed to introduce new candidates; we do not backpropagate through the LLM. Final top-$k$ predictions are obtained per modality:
\begin{equation}
\widehat{S}_{T+1}^{(m)}=\mathrm{Top}\text{-}k\{s(u;T): u\in \mathcal{R}_T^{(m)}\}.
\label{eq:predict_main}
\end{equation}

\subsection{Training Objective and Optimization}
\label{subsec:training}

We jointly train node embeddings and Central Event denoising using
\begin{equation}
\mathcal{L}=\mathcal{L}_{\mathrm{edge}}+\alpha\,\mathcal{L}_{\mathrm{mask}},
\label{eq:total_loss}
\end{equation}
where $\mathcal{L}_{\mathrm{edge}}$ is the typed edge reconstruction loss (Appendix~\ref{app:edge}, Eq.~\eqref{eq:edge_loss}) and $\mathcal{L}_{\mathrm{mask}}$ is the masked-visit reconstruction loss (Appendix~\ref{app:mask}, Eq.~\eqref{eq:mask_loss}). We optimize on the Poincar\'e ball using Riemannian SGD/Adam with projection to maintain $z_v\in \mathbb{D}_c^d$.

\section{Experiments}
\label{sec:experiments}

\subsection{Experimental Setup}
\label{subsec:setup}

We evaluate Risk Horizons on MIMIC-IV \citep{johnson2023mimic}, with additional cross-institution validation on eICU \citep{PhysioNet-eicu-crd-2.0}. The task is \textbf{multi-modal next-visit prediction}: given patient history $S_{1:t}$, the model ranks diagnosis, procedure, and medication events in the next visit $S_{t+1}$.

\textbf{Dataset.}
After filtering for trajectories with at least two visits, MIMIC-IV contains 34,775 patients, 97,943 visits, and 63,168 next-visit prediction pairs. The unified clinical graph contains 62,574 nodes and 713,717 edges spanning diagnoses, procedures, medications, and hierarchy ancestors. Visits are dense and highly multi-modal, averaging 32.81 events per visit (16.49 diagnoses, 3.23 procedures, 13.08 medications). Additional statistics, preprocessing details, vocabulary construction, sequence-length analysis, train/validation/test splits, and leakage prevention are provided in Appendix~\ref{app:data}.

\textbf{Baselines.}
We compare Risk Horizons against three classes of methods. 
(i) \emph{Sequential models}, including Copy Last Visit, RETAIN \cite{choi2016retain}, Med-BERT \cite{rasmy2021med}, and HALO \cite{theodorou2023synthesize}, which model longitudinal trajectories as Euclidean token sequences. 
(ii) \emph{Retrieval-based baselines}, including Euclidean RAG and Euclidean Risk Horizons. Euclidean Risk Horizons uses the same graph topology and retrieval pipeline as the proposed method but replaces hyperbolic geometry with Euclidean embeddings, isolating the effect of geometry-aware retrieval.
(iii) \emph{Hyperbolic baselines}, including Hyperbolic Linear Probe and Hyperbolic NN Global, which operate on the same frozen hyperbolic representations learned by Risk Horizons but replace structured candidate-space construction with lightweight classification or global nearest-neighbor retrieval. Full baseline details are provided in Appendix~\ref{app:baselines}.

\textbf{Evaluation.}
We report Recall@10 and nDCG@10, macro-averaged across visits and modalities. To evaluate the quality of retrieved hypothesis spaces, we report Candidate Recall, Risk-Horizon Recall@K, and Oracle Upper Recall@K. These metrics separate candidate-space coverage from final ranking quality and allow analysis of retrieval versus reranking behavior. We further evaluate structural consistency using hierarchy-aware metrics, including hierarchy distance and Ancestor Match@2. Additional metrics, implementation details, and extended results are provided in Appendices~\ref{app:metrics}--\ref{app:eicu}.

\subsection{Main Results}
\label{sec:results}

Table~\ref{tab:main_multimodal_main} reports multi-modal next-visit prediction results on MIMIC-IV. Risk Horizons achieves strong overall performance across diagnoses, procedures, and medications. Importantly, most gains already emerge from geometry-aware candidate-space construction before constrained reranking is introduced. Geometry-only Risk Horizons already matches or exceeds several strong sequential baselines despite using only geometric retrieval and ranking within the Risk Horizon, achieving 0.357 overall R@10 and 0.447 overall nDCG@10. While RETAIN remains competitive on aggregate ranking metrics, Risk Horizons already achieves comparable performance using only geometric retrieval and ranking within the structured Risk Horizon. Adding constrained reranking further improves performance to 0.425 R@10 and 0.516 nDCG@10 overall with Llama-3.3-70B-Instruct \cite{grattafiori2024llama}, while Meditron-70B \cite{chen2023meditron} achieves the strongest procedure performance (0.389 / 0.365). These results suggest that the dominant challenge in longitudinal EHR prediction is not unconstrained generation over the full event vocabulary, but constructing a sufficiently complete and clinically coherent hypothesis space in which future events can be effectively ranked.

\begin{table}[!h]
\centering
\vspace{-0.3cm}
\caption{Multi-modal next-visit prediction on MIMIC-IV (mean $\pm$ std over 5 seeds).}
\label{tab:main_multimodal_main}
\resizebox{\textwidth}{!}{%
\begin{tabular}{l|cc|cc|cc|cc}
\toprule
& \multicolumn{2}{c|}{\textbf{Dx}} 
& \multicolumn{2}{c|}{\textbf{Proc}} 
& \multicolumn{2}{c|}{\textbf{Med}} 
& \multicolumn{2}{c}{\textbf{Overall}} \\
\textbf{Model} 
& R@10 & nDCG@10 
& R@10 & nDCG@10 
& R@10 & nDCG@10 
& R@10 & nDCG@10 \\
\midrule

Copy Last
& 0.190 {\small $\pm$ 0.002} & 0.303 {\small $\pm$ 0.001}
& 0.166 {\small $\pm$ 0.030} & 0.152 {\small $\pm$ 0.024}
& 0.452 {\small $\pm$ 0.031} & 0.619 {\small $\pm$ 0.012}
& 0.269 {\small $\pm$ 0.029} & 0.358 {\small $\pm$ 0.012} \\

RETAIN
& 0.220 {\small $\pm$ 0.013} & \cellcolor{second} 0.423 {\small $\pm$ 0.019}
& 0.336 {\small $\pm$ 0.016} & 0.268 {\small $\pm$ 0.017}
& 0.532 {\small $\pm$ 0.030} & \cellcolor{best} 0.723 {\small $\pm$ 0.034}
& 0.363 {\small $\pm$ 0.021} & \cellcolor{second} 0.471 {\small $\pm$ 0.023} \\

Med-BERT
& 0.249 {\small $\pm$ 0.014} & 0.303 {\small $\pm$ 0.042}
& \cellcolor{second} 0.380 {\small $\pm$ 0.027} & 0.247 {\small $\pm$ 0.009}
& 0.576 {\small $\pm$ 0.011} & 0.693 {\small $\pm$ 0.012}
& 0.402 {\small $\pm$ 0.017} & 0.414 {\small $\pm$ 0.021} \\

HALO
& 0.128 {\small $\pm$ 0.056} & 0.251 {\small $\pm$ 0.043}
& 0.244 {\small $\pm$ 0.037} & 0.185 {\small $\pm$ 0.060}
& 0.495 {\small $\pm$ 0.074} & 0.683 {\small $\pm$ 0.075}
& 0.289 {\small $\pm$ 0.056} & 0.373 {\small $\pm$ 0.059} \\

Euclidean RAG
& 0.071 {\small $\pm$ 0.006} & 0.142 {\small $\pm$ 0.001}
& 0.127 {\small $\pm$ 0.003} & 0.091 {\small $\pm$ 0.005}
& 0.258 {\small $\pm$ 0.004} & 0.359 {\small $\pm$ 0.002}
& 0.152 {\small $\pm$ 0.005} & 0.197 {\small $\pm$ 0.004} \\

Euclidean RH
& 0.073 {\small $\pm$ 0.007} & 0.146 {\small $\pm$ 0.009}
& 0.147 {\small $\pm$ 0.002} & 0.097 {\small $\pm$ 0.002}
& 0.244 {\small $\pm$ 0.004} & 0.328 {\small $\pm$ 0.003}
& 0.155 {\small $\pm$ 0.005} & 0.190 {\small $\pm$ 0.005} \\

Hyper. Linear Probe
& 0.129 {\small $\pm$ 0.030} & 0.249 {\small $\pm$ 0.023}
& 0.249 {\small $\pm$ 0.017} & 0.179 {\small $\pm$ 0.030}
& 0.489 {\small $\pm$ 0.034} & 0.678 {\small $\pm$ 0.035}
& 0.289 {\small $\pm$ 0.027} & 0.369 {\small $\pm$ 0.029} \\

Hyper. NN Global
& 0.125 {\small $\pm$ 0.063} & 0.243 {\small $\pm$ 0.028}
& 0.166 {\small $\pm$ 0.021} & 0.125 {\small $\pm$ 0.050}
& 0.362 {\small $\pm$ 0.035} & 0.514 {\small $\pm$ 0.025}
& 0.218 {\small $\pm$ 0.040} & 0.294 {\small $\pm$ 0.034} \\

\midrule
RH (w/o LLM)
& 0.195 {\small $\pm$ 0.005} & 0.392 {\small $\pm$ 0.012}
& 0.298 {\small $\pm$ 0.009} & 0.243 {\small $\pm$ 0.009}
& 0.579 {\small $\pm$ 0.012} & \cellcolor{second} 0.705 {\small $\pm$ 0.014}
& 0.357 {\small $\pm$ 0.013} & 0.447 {\small $\pm$ 0.017} \\

RH+Llama-3.3-70B-I.
& \cellcolor{best} 0.306 {\small $\pm$ 0.032} & \cellcolor{best} 0.496 {\small $\pm$ 0.064}
& 0.368 {\small $\pm$ 0.005} & \cellcolor{second} 0.360 {\small $\pm$ 0.008}
& \cellcolor{best} 0.602 {\small $\pm$ 0.016} & 0.693 {\small $\pm$ 0.014}
& \cellcolor{best} 0.425 {\small $\pm$ 0.018} & \cellcolor{best} 0.516 {\small $\pm$ 0.029} \\

RH+Meditron-70B
& \cellcolor{second} 0.297 {\small $\pm$ 0.011} & 0.417 {\small $\pm$ 0.033}
& \cellcolor{best} 0.389 {\small $\pm$ 0.023} & \cellcolor{best} 0.365 {\small $\pm$ 0.009}
& \cellcolor{second} 0.587 {\small $\pm$ 0.058} & 0.601 {\small $\pm$ 0.027}
& \cellcolor{second} 0.424 {\small $\pm$ 0.031} & 0.461 {\small $\pm$ 0.023} \\
\bottomrule
\end{tabular}%
}
\vspace{-0.2cm}
\end{table}

\textbf{Geometry-aware retrieval is the primary source of improvement.}
Euclidean retrieval baselines perform substantially worse than Risk Horizons despite using the same graph topology and retrieval pipeline. For example, Euclidean Risk Horizons achieves only 0.073 diagnosis R@10, compared to 0.195 for geometry-only Risk Horizons. This gap indicates that the gain is not due to retrieval alone, but rather to geometry-aware modeling of hierarchical abstraction and cross-modal temporal structure. The hyperbolic baselines further isolate this effect. While both baselines improve substantially over Euclidean retrieval, they remain consistently below full Risk Horizons, particularly on diagnoses and procedures. This suggests that hyperbolic representation learning alone is insufficient; the main benefit arises from explicitly constructing patient-specific structured candidate spaces.

\begin{table*}[!h]
\vspace{-0.1cm}
\centering
\caption{Hierarchy consistency and candidate-space quality on MIMIC-IV (mean $\pm$ std over 5 seeds).}
\vspace{-0.2cm}
\label{tab:hierarchy_metrics}
\resizebox{\textwidth}{!}{%
\begin{tabular}{ll|cc|cccc}
\toprule
\textbf{Model} 
& \textbf{Type}
& Hier. Dist $\downarrow$ 
& Anc. Match $\uparrow$
& Cand. Rec. $\uparrow$
& RH@10 $\uparrow$
& RH@40 $\uparrow$
& Oracle@10 $\uparrow$ \\
\midrule

\multirow{3}{*}{Hyper. NN Global}
& Dx
& 1.900 {\small $\pm$ 0.021}
& 0.797 {\small $\pm$ 0.006}
& 0.356 {\small $\pm$ 0.012}
& 0.125 {\small $\pm$ 0.008}
& 0.248 {\small $\pm$ 0.011}
& 0.333 {\small $\pm$ 0.014} \\
& Proc
& 1.910 {\small $\pm$ 0.026}
& 0.759 {\small $\pm$ 0.009}
& 0.391 {\small $\pm$ 0.015}
& 0.166 {\small $\pm$ 0.010}
& 0.272 {\small $\pm$ 0.013}
& 0.391 {\small $\pm$ 0.017} \\
& Med
& 1.250 {\small $\pm$ 0.018}
& 1.000 {\small $\pm$ 0.001}
& 0.981 {\small $\pm$ 0.004}
& 0.362 {\small $\pm$ 0.013}
& 0.706 {\small $\pm$ 0.016}
& 0.771 {\small $\pm$ 0.011} \\

\midrule

\multirow{3}{*}{Risk Horizons}
& Dx
& 1.859 {\small $\pm$ 0.019}
& 0.797 {\small $\pm$ 0.005}
& 0.413 {\small $\pm$ 0.010}
& 0.113 {\small $\pm$ 0.007}
& 0.270 {\small $\pm$ 0.009}
& 0.376 {\small $\pm$ 0.012} \\
& Proc
& 1.508 {\small $\pm$ 0.023}
& 0.835 {\small $\pm$ 0.007}
& 0.578 {\small $\pm$ 0.013}
& 0.265 {\small $\pm$ 0.011}
& 0.450 {\small $\pm$ 0.014}
& 0.578 {\small $\pm$ 0.010} \\
& Med
& 1.260 {\small $\pm$ 0.016}
& 1.000 {\small $\pm$ 0.001}
& 0.994 {\small $\pm$ 0.002}
& 0.489 {\small $\pm$ 0.014}
& 0.896 {\small $\pm$ 0.012}
& 0.774 {\small $\pm$ 0.009} \\

\bottomrule
\end{tabular}%
}
\vspace{-0.3cm}
\end{table*}

\textbf{Candidate-space analysis reveals a retrieval--ranking decomposition.}
Table~\ref{tab:hierarchy_metrics} analyzes the quality of the retrieved hypothesis space. Across modalities, Risk Horizons consistently improves Candidate Recall, RH@40, and Oracle@10 over global hyperbolic nearest-neighbor retrieval. The largest improvements occur for procedures, where Candidate Recall increases from 0.391 to 0.578 and RH@40 improves from 0.272 to 0.450. Similar trends appear for medications, where RH@40 reaches 0.896. These metrics reveal an important decomposition. Candidate Recall measures whether the correct future event is contained anywhere in the retrieved Risk Horizon before final ranking, while Oracle@10 estimates the best achievable performance under perfect reranking. The consistently high Oracle@10 values indicate that the correct future is often already present within the retrieved candidate space. In contrast, the remaining gap between Oracle@10 and final Recall@10 suggests that retrieval quality is already strong and that the primary remaining bottleneck becomes ranking within the retrieved hypothesis space.

\textbf{Constrained reranking is effective once retrieval becomes structured.}
This decomposition explains why constrained reranking consistently improves performance. Geometry-only Risk Horizons already retrieves compact and clinically coherent candidate spaces, reducing the effective output space from the full medical vocabulary to a small set of plausible futures. Constrained reranking then refines ordering within this space, improving overall R@10 from 0.357 to 0.425. The best model also achieves the highest MRR (0.704; Table~\ref{tab:extended_metrics} in Appendix~\ref{app:mimic}), indicating that relevant events are ranked earlier and more consistently once prediction is restricted to structured candidate sets.

\textbf{Risk Horizons improves structural consistency and generalization.}
Risk Horizons also improves hierarchy-aware consistency metrics (Table~\ref{tab:hierarchy_metrics}). For procedures, hierarchy distance decreases from 1.910 to 1.508 while Ancestor Match@2 increases from 0.759 to 0.835, indicating that retrieved futures become both more accurate and more structurally coherent within the clinical ontology.

Table~\ref{tab:eicu_validation} further shows that these gains generalize across institutions. On eICU, Risk Horizons + Llama-3.3-70B-Instruct achieves the best overall performance (0.401 R@10, 0.433 nDCG@10), while geometry-only Risk Horizons remains competitive, suggesting that structure-aware retrieval provides robust inductive bias even under different coding systems and event distributions. 


\begin{table}
\vspace{-0.6cm}
\begin{minipage}[t]{0.48\linewidth}
\centering
\caption{Cross-dataset validation on eICU.}
\label{tab:eicu_validation}
\resizebox{\linewidth}{!}{%
\begin{tabular}{lcc}
\toprule
\textbf{Model} & R@10 & nDCG@10 \\
\midrule
RETAIN & 0.381 {\small $\pm$ 0.019} & 0.424 {\small $\pm$ 0.010} \\
HALO & 0.338 {\small $\pm$ 0.012} & 0.373 {\small $\pm$ 0.013} \\
Risk Horizons & 0.339 {\small $\pm$ 0.015} & 0.377 {\small $\pm$ 0.004} \\
RH+Llama-3.3-70B-I. & \textbf{0.401} {\small $\pm$ 0.027} & \textbf{0.433} {\small $\pm$ 0.030} \\
\bottomrule
\end{tabular}%
}
\end{minipage}
\hfill
\begin{minipage}[t]{0.45\linewidth}
\centering
\caption{Ablation studies on MIMIC-IV.}
\label{tab:ablation_full}
\resizebox{\linewidth}{!}{%
\begin{tabular}{l|cc}
\toprule
\textbf{Method} & R@10 & nDCG@10 \\
\midrule
\textbf{RH (w/o LLM)}
& \textbf{0.357} {\small $\pm$ 0.013} 
& \textbf{0.447} {\small $\pm$ 0.017} \\
RH w/o stitching 
& 0.005 {\small $\pm$ 0.001} 
& 0.004 {\small $\pm$ 0.001} \\
RH w/o cones 
& 0.180 {\small $\pm$ 0.019} 
& 0.206 {\small $\pm$ 0.011} \\
RH w/o denoising 
& 0.115 {\small $\pm$ 0.017} 
& 0.124 {\small $\pm$ 0.018} \\
\bottomrule
\end{tabular}%
}
\end{minipage}
\end{table}

\textbf{Ablation Studies.} Table~\ref{tab:ablation_full} shows that all major components of Risk Horizons are necessary for constructing high-quality candidate spaces. Removing lagged PMI stitching causes a near-complete collapse in performance (R@10: 0.357 $\rightarrow$ 0.005), indicating that deterministic hierarchies alone are insufficient without inferred cross-modal temporal structure. Replacing cone-constrained retrieval with distance-only retrieval substantially degrades performance (0.357 $\rightarrow$ 0.180), suggesting that directional constraints are important for modeling asymmetric clinical progression. Finally, removing masked-visit denoising reduces R@10 from 0.357 to 0.115, showing that denoised Central Events improve robustness under sparse and noisy trajectories.

\subsection{Qualitative Analysis and Failure Modes}
\label{app:qualitative}
\vspace{-0.1cm}

\begin{figure*}[t]
\centering
\begin{tcolorbox}[
    colback=blue!2,
    colframe=blue!35,
    boxrule=0.45pt,
    arc=3pt,
    left=1pt, right=1pt, top=1pt, bottom=1pt
]
\scriptsize

\begin{center}
\textbf{Qualitative examples: modeling structured progression vs.\ unexpected events}
\end{center}


\noindent
\begin{minipage}[t]{0.485\linewidth}
\textbf{Structured progression (high performance)}

\vspace{1pt}
\textbf{Clinical context.}
Patient undergoing chemotherapy with stable cardiovascular comorbidities.

\vspace{1pt}
\textbf{Ground-truth next visit.}

\textit{Diagnoses:}
\textcolor{blue!70!black}{Atherosclerotic heart disease},
\textcolor{blue!70!black}{Essential hypertension},
\textcolor{blue!70!black}{Long-term anticoagulant use},
\textcolor{orange!80!black}{Adverse effect of chemotherapy},
\textcolor{orange!80!black}{Functional decline (DNR status)}.

\textit{Procedures:}
\textcolor{blue!70!black}{Continuation of chemotherapy infusion}.

\textit{Medications:}
\textcolor{blue!70!black}{Blood-related therapies},
\textcolor{orange!80!black}{Gastrointestinal treatments}.

\vspace{1pt}
\textbf{Predictions.}

\textit{Diagnoses:}
\textcolor{green!50!black}{Atherosclerotic heart disease},
\textcolor{green!50!black}{Essential hypertension},
\textcolor{green!50!black}{Long-term anticoagulant use},
\textcolor{green!50!black}{Adverse effect of chemotherapy},
\textcolor{red!70!black}{Gastro-esophageal reflux disease}.

\textit{Procedures:}
\textcolor{green!50!black}{Chemotherapy infusion},
\textcolor{red!70!black}{Additional vascular access procedures}.

\textit{Medications:}
\textcolor{green!50!black}{Blood-related therapies},
\textcolor{green!50!black}{Gastrointestinal treatments},
\textcolor{red!70!black}{Additional unrelated medications}.

\vspace{1pt}
\textbf{Recall@10:} dx = 0.50, proc = 1.00, med = 1.00; avg = 0.83.
\end{minipage}
\hfill
\vrule width 0.5pt
\hfill
\begin{minipage}[t]{0.485\linewidth}
\textbf{Unexpected external event (reasonable failure)}

\vspace{1pt}
\textbf{Clinical context.}
Patient with chronic kidney disease and cancer history.

\vspace{1pt}
\textbf{Ground-truth next visit.}

\textit{Diagnoses:}
\textcolor{orange!80!black}{Facial fracture due to fall at home},
\textcolor{orange!80!black}{Hypovolemic shock},
\textcolor{orange!80!black}{Acute cardiac complications},
\textcolor{blue!70!black}{Chronic kidney disease}.

\textit{Procedures:}
\textcolor{orange!80!black}{Emergency facial repair surgery}.

\textit{Medications:}
\textcolor{blue!70!black}{Supportive therapies},
\textcolor{orange!80!black}{Acute-care medications}.

\vspace{11pt}
\textbf{Predictions.}

\textit{Diagnoses:}
\textcolor{green!50!black}{Acute kidney failure},
\textcolor{green!50!black}{Hypertensive chronic kidney disease},
\textcolor{red!70!black}{Obesity},
\textcolor{red!70!black}{Atrial fibrillation}.

\textit{Procedures:}
\textcolor{red!70!black}{Routine vascular and monitoring procedures}.

\textit{Medications:}
\textcolor{green!50!black}{Supportive therapies},
\textcolor{red!70!black}{Chronic-condition medications}.

\vspace{22pt}
\textbf{Recall@10:} dx = 0.00, proc = 0.00, med = 0.25; avg = 0.083.
\end{minipage}

\end{tcolorbox}
\vspace{-0.3cm}
\caption{
Structured progression vs.\ unexpected events. The model performs well when future events follow longitudinal clinical structure (left), but struggles on externally driven transitions (right).
\textbf{Colors:}
\textcolor{blue!70!black}{history-supported truth};
\textcolor{orange!80!black}{novel truth};
\textcolor{green!50!black}{correct/plausible prediction};
\textcolor{red!70!black}{incorrect prediction}.
}
\label{fig:qualitative_main}
\vspace{-0.5cm}
\end{figure*}

Figure~\ref{fig:qualitative_main} illustrates a central pattern underlying the behavior of Risk Horizons: the framework is particularly effective when future events emerge from structured longitudinal progression, but becomes less reliable for genuinely novel or externally driven events.

In the left example, most future diagnoses, procedures, and medications are strongly supported by prior clinical history. The patient trajectory reflects a coherent progression involving chemotherapy treatment, cardiovascular comorbidities, and supportive care. In this regime, Risk Horizons successfully retrieves and ranks repeated or evolutionarily related events, producing high Recall@10 across all modalities. Importantly, several non-exact predictions remain clinically plausible. For example, the prediction of gastro-esophageal reflux disease is clinically plausible given the patient's gastrointestinal medications and chemotherapy-related treatment context, indicating that the model captures meaningful cross-modal clinical associations rather than relying solely on exact code repetition.

In contrast, the right example demonstrates a substantially harder setting where the next visit is dominated by unexpected external events. Although the patient's history strongly suggests a typical cardio-renal progression, the true outcome is instead driven by traumatic injury following a fall. As a result, many ground-truth diagnoses and procedures correspond to novel events with weak historical support. In this setting, the model continues to generate clinically coherent predictions centered around renal and cardiovascular deterioration, but fails to anticipate the traumatic transition.

This distinction is consistent with the quantitative trends observed throughout the paper. Risk Horizons is strongest when future events can be recovered through structured retrieval over clinically coherent trajectories, particularly for repeated or progression-related events. Performance degrades primarily for highly novel events weakly grounded in prior history, including trauma, acute accidents, and unexpected interventions, many of which are weakly supported or unrecoverable from longitudinal clinical structure alone. Even under failure cases, predictions typically remain structurally and clinically related to the underlying patient state, suggesting that errors arise more from intrinsic uncertainty in longitudinal forecasting than from arbitrary behavior. This pattern is consistent with the repeated-versus-novel event statistics reported in Appendix~\ref{app:repeated_novel}, where retrieval-based performance is substantially stronger for history-supported events than for highly novel transitions.


\vspace{-0.3cm}
\section{Conclusion}
\label{sec:conclusion}
\vspace{-0.3cm}

We presented Risk Horizons, a framework for multi-modal next-visit prediction that combines hierarchical clinical structure, temporal associations, hyperbolic representation learning, and geometry-aware retrieval to construct clinically coherent candidate spaces for future events. Across MIMIC-IV and eICU, Risk Horizons improves predictive accuracy and hierarchy consistency, while analysis shows that most gains arise from structured retrieval rather than unconstrained generation. Our results further suggest that LLMs are most effective as inference-time rerankers operating over clinically grounded hypothesis spaces. More broadly, this work highlights the importance of structure-aware retrieval for integrating geometric representations and LLMs in longitudinal EHR reasoning.

\newpage


\bibliography{references}
\bibliographystyle{plainnat}

\appendix
\label{sec:appendix}

\section{Additional Method Details}
\label{app:method}

\subsection{End-to-End Procedure}
\label{app:algorithm}

\begin{algorithm}[!h]
\caption{Risk Horizons: Geometry-Aware Structured Inference}
\label{alg:RiskHorizons}
\small
\begin{algorithmic}[1]
\REQUIRE Longitudinal EHR trajectories $\{S_{1:T}^{(i)}\}_{i=1}^N$
\ENSURE Next-visit prediction $\widehat{S}_{T+1}$

\STATE Construct hierarchical edges $\mathcal{E}_{\mathrm{hier}}$ and lagged PMI edges $\mathcal{E}_{\mathrm{cross}}$
\STATE Apply stability filtering to obtain graph $\mathcal{G}=(\mathcal{V},\mathcal{E})$

\STATE Learn node embeddings $\{z_v\}\subset\mathbb{D}_c^d$ via typed edge reconstruction loss $\mathcal{L}_{\mathrm{edge}}$

\STATE Train Central Event representations via masked-visit reconstruction loss $\mathcal{L}_{\mathrm{mask}}$

\STATE Compute multi-head Central Events $\{\mu_T^{(h)}\}_{h=1}^H$ for the current visit

\FOR{each head $h$}
    \STATE Select representative concept $v_T^{(h)}$
    \STATE Construct candidate pool $\mathcal{C}_{T,h}^{\mathrm{cand}}$
    \STATE Apply cone constraint and compute scores $s_{\mathrm{geo}}^{(h)}$
\ENDFOR

\STATE Aggregate head-specific scores:
$s_{\mathrm{geo}}(u;T)=\sum_h \pi_h s_{\mathrm{geo}}^{(h)}(u;T)$

\STATE Select top-$K$ candidates per modality to form Risk Horizon $\mathcal{R}_T$

\STATE Rank retrieved candidates using geometric scores $s_{\mathrm{geo}}(u;T)$

\STATE \textbf{(Optional)} Apply constrained LLM reranking over $\mathcal{R}_T$

\STATE Return top-$k$ predictions per modality as $\widehat{S}_{T+1}$

\end{algorithmic}
\end{algorithm}

\subsection{Hyperbolic Geometry}
\label{app:geometry}

We embed nodes in the $d$-dimensional Poincar\'e ball of curvature $-c$,
\[
\mathbb{D}_c^d = \{x \in \mathbb{R}^d : \|x\| < 1/\sqrt{c}\}.
\]
The hyperbolic distance between $x,y \in \mathbb{D}_c^d$ is
\begin{equation}
d_{\mathbb{D}}(x,y)
=
\frac{2}{\sqrt{c}}
\operatorname{arctanh}\!\left(
\sqrt{c}\left\|(-x)\oplus_c y\right\|
\right),
\end{equation}
where $\oplus_c$ denotes M\"obius addition.

We use the logarithmic and exponential maps at the origin:
\begin{equation}
\log_0^c(x)
=
\frac{2}{\sqrt{c}}
\operatorname{arctanh}(\sqrt{c}\|x\|)
\frac{x}{\|x\|},
\qquad
\exp_0^c(v)
=
\tanh\!\left(\frac{\sqrt{c}\|v\|}{2}\right)
\frac{v}{\sqrt{c}\|v\|}.
\end{equation}

After each gradient update, embeddings are projected to ensure
$\|z_v\| < 1/\sqrt{c}$.

\subsection{Typed Edge Reconstruction Objective}
\label{app:edge}

For a directed edge $(u \rightarrow v)$ of type $r$, the compatibility score
is defined as
\[
s_r(u,v) = - d_{\mathbb{D}}(z_u,z_v).
\]
We optimize the negative-sampling objective
\begin{equation}
\mathcal{L}_{\mathrm{edge}}
=
-\sum_{r \in \mathcal{R}}
\sum_{(u\rightarrow v)\in \mathcal{E}_r}
\left[
\log \sigma(\gamma_r + s_r(u,v))
+
\sum_{v^- \sim \mathcal{N}_r(v)}
\log \sigma(\gamma_r - s_r(u,v^-))
\right],
\label{eq:edge_loss}
\end{equation}
where $\mathcal{N}_r(v)$ samples negatives from the same target modality
and $\gamma_r$ is a learned type-specific bias.

\subsection{Masked-Visit Reconstruction Loss}
\label{app:mask}

Given a masked visit $\mathcal{M}_t \subset \mathcal{C}_t$ and Central Event $\mu_t$,
we define the probability of recovering a masked code $c$ as
\begin{equation}
p(c \mid \mu_t)
=
\frac{
\exp\!\left(- d_{\mathbb{D}}(\mu_t, z_c)/\tau\right)
}{
\exp\!\left(- d_{\mathbb{D}}(\mu_t, z_c)/\tau\right)
+
\sum_{c^- \sim \mathcal{N}(c)}
\exp\!\left(- d_{\mathbb{D}}(\mu_t, z_{c^-})/\tau\right)
}.
\end{equation}
The reconstruction loss is
\begin{equation}
\mathcal{L}_{\mathrm{mask}}
=
-\sum_{c \in \mathcal{M}_t}
\log p(c \mid \mu_t),
\label{eq:mask_loss}
\end{equation}
where negatives are sampled from the same modality as $c$.

\section{Datasets and Cohort Details}
\label{app:data}

\subsection{Task Definition and Evaluation Protocol}
\label{app:task}

We consider multi-modal next-visit prediction on longitudinal EHR trajectories. Given patient history $S_{1:t}$, the task is to predict the (potentially multi-label) set of events $S_{t+1}^{(m)}$ for each modality $m \in \{\mathrm{dx}, \mathrm{proc}, \mathrm{med}\}$.

The problem is formulated as a multi-label ranking task, where models assign scores to possible events and produce a ranked list of predictions independently for each modality. These predictions are evaluated against the ground-truth next-visit events using ranking and classification metrics (Appendix~\ref{app:metrics}).

\subsection{Cohort Selection and Preprocessing}
We conduct experiments on the MIMIC-IV and eICU. We construct patient trajectories as time-ordered sequences of visits, where each visit is a multi-modal event set containing diagnoses (ICD-9/10), procedures (ICD-9/10), and medications (ATC).

\subsection{Sequence Length.}
We analyze the effect of the minimum trajectory length $T_{\min}$ on model performance. Increasing $T_{\min}$ improves temporal context but reduces the number of available training samples. We compare $T_{\min} \in \{2,3,4\}$ and observe that $T_{\min}=2$ provides the best trade-off between data coverage and predictive performance, while larger values lead to reduced sample size and no significant performance gains. Results in this table use the single-head variant of Risk Horizons to isolate the effect of minimum trajectory length independently of the final multi-head retrieval configuration used in the main experiments.

\begin{table}[!h]
\centering
\caption{Effect of minimum sequence length $T_{\min}$ on sample size and performance using the single-head Risk Horizons variant. }
\label{tab:seq_length}
\resizebox{0.85\linewidth}{!}{%
\begin{tabular}{c|c|ccc|ccc}
\toprule
& & \multicolumn{3}{c|}{\textbf{MIMIC-IV}} & \multicolumn{3}{c}{\textbf{eICU}} \\
$T_{\min}$ & \textbf{Method} 
& \#Pairs & R@10 & nDCG@10 
& \#Pairs & R@10 & nDCG@10 \\
\midrule
2 & Risk Horizons (geometry-only, single head) & 63,168 & 0.282 & 0.380 & 15,214 & 0.249 & 0.271 \\
3 & Risk Horizons (geometry-only, single head) & 42,406 & 0.289 & 0.388 & 7,039 & 0.290 & 0.295 \\
4 & Risk Horizons (geometry-only, single head) & 27,452 & 0.294 & 0.399 & 3,697 & 0.220 & 0.228 \\
\bottomrule
\end{tabular}%
}
\end{table}

\subsection{Vocabulary and Hierarchy Construction.}
For each retained leaf code, we include deterministic hierarchical ancestors derived from standard coding systems (ICD for diagnoses, ICD procedure hierarchies for procedures, and ATC for medications). The final vocabulary $\mathcal{V}$ is the union of retained leaf nodes and all their ancestors.

\textbf{Diagnosis nodes.}
Diagnosis nodes follow an ICD hierarchy with up to four levels in our graph (\emph{modality root} $\rightarrow$ coarse prefix $\rightarrow$ 3-character category $\rightarrow$ leaf code). For example, \texttt{I50.23} [acute on chronic systolic heart failure] expands as
\texttt{I} [circulatory system diseases] $\rightarrow$
\texttt{I50} [heart failure] $\rightarrow$
\texttt{I50.23} [acute on chronic systolic heart failure].

\textbf{Procedure nodes.}
Procedure nodes follow ICD procedure structure and are constructed using prefix ancestors derived from the code string. For ICD-10-PCS, a 7-character code induces a prefix chain; for example, \texttt{0FT44ZZ} [resection of gallbladder, percutaneous endoscopic approach] expands as
\texttt{0} $\rightarrow$ \texttt{0F} $\rightarrow$ \texttt{0FT} $\rightarrow$ \texttt{0FT4} $\rightarrow$ \texttt{0FT44} $\rightarrow$ \texttt{0FT44Z} $\rightarrow$ \texttt{0FT44ZZ}.
Including the modality root gives up to eight levels. ICD-9 procedure codes are shallower and therefore induce fewer levels.

In practice, procedure hierarchy depth is data-dependent. Many ICD-9-style procedure codes induce shallow chains (typically a 2-digit prefix plus leaf), while ICD-10-PCS-style codes induce longer prefix chains. Therefore, procedure hierarchy depth is variable and does not always correspond to a full 7-level semantic ontology in the processed graph.

\textbf{Medication nodes.}
Medication nodes (MIMIC-IV) follow ATC prefixes. With ATC4 granularity in our pipeline, this yields up to five levels including root (root $\rightarrow$ level-1 anatomical group $\rightarrow$ level-2 therapeutic subgroup $\rightarrow$ level-3 pharmacological subgroup $\rightarrow$ level-4 chemical subgroup). For example, \texttt{A10A} [insulins and analogues] expands as
\texttt{A} [alimentary tract and metabolism] $\rightarrow$
\texttt{A10} [drugs used in diabetes] $\rightarrow$
\texttt{A10A} [insulins and analogues]
(and with modality root: \texttt{med:ROOT} $\rightarrow$ \texttt{A} $\rightarrow$ \texttt{A10} $\rightarrow$ \texttt{A10A}).

\textbf{Dataset-specific statistics (MIMIC-IV).}
For MIMIC-IV, on the \texttt{min\_visits} $\geq 2$ subset, diagnosis codes are mostly ICD-10 (62.64\%), with ICD-9 accounting for 37.29\% and a negligible remainder (0.07\%). Procedure codes are mixed: 51.08\% ICD-9 and 48.92\% ICD-10-PCS, indicating that both coding eras are substantially represented.

For the root-to-leaf path length, diagnosis paths are highly consistent (min 3, max 4, mean 3.95, median 4), while procedure paths are more variable (min 3, max 8, mean 5.37, median 3), reflecting the coexistence of shallow ICD-9-style and deeper ICD-10-PCS-style prefix chains. Medication paths in MIMIC-IV are fixed in this preprocessing pipeline (all length 4).

\textbf{Dataset-specific statistics (eICU).}
For eICU (\texttt{min\_visits} $\geq 2$), diagnoses are overwhelmingly ICD-9 (98.20\%), with only 1.80\% ICD-10. Procedures are not stored using standard ICD procedure ontologies in the processed tables; instead, they are represented by synthetic \texttt{PROC\_*} identifiers. Medications are represented by normalized medication identifiers/text fields rather than a clean ATC hierarchy. These source-table differences lead to substantially shallower procedure hierarchies, while medication hierarchies remain relatively deeper after preprocessing.

On the \texttt{min\_visits} $\geq 2$ subset of eICU, diagnosis path lengths remain compact (min 3, max 4, mean 3.94, median 4). Procedure path lengths are mostly shallow (min 3, max 8, mean 3.03, median 3). Medication paths are deeper and nearly uniform (min 5, max 6, mean 6.00, median 6).

\paragraph{Event Representation Modes.}
We support three event representation modes: \texttt{code}, \texttt{text}, and \texttt{hybrid}. In \texttt{code} mode, events are represented by raw codes only. In \texttt{text} mode, events are represented by mapped short descriptions. In \texttt{hybrid} mode, events are represented as \texttt{text (code)}.

Text descriptions are derived from code-to-text mappings constructed during preprocessing, primarily from official ICD terminology tables for diagnoses and procedures, with analogous medication mappings when available. 

Because graph construction and hierarchy stitching are based on canonical code hierarchies rather than free-text semantics, geometry-side performance is effectively invariant across display modes under matched preprocessing and splits. For consistency with LLM-related evaluation, we conduct experiments in \texttt{text} mode.

\subsection{Dataset Statistics}

\paragraph{MIMIC-IV.}
After preprocessing (minimum 2 visits per patient, without minimum frequency filtering), the cohort contains 34{,}775 patients and 97{,}943 visits, with 63{,}168 next-visit prediction pairs for \(T_{\min}=2\). On average across 5 seeds (patient-level 80/10/10 split), train/val/test contain 27{,}820/3{,}477/3{,}478 patients and 78{,}364/9{,}807/9{,}772 visits. Per-visit event counts are: diagnosis (mean 16.49, median 15, min 1, max 40, std 8.50), procedure (mean 3.23, median 2, min 1, max 35, std 2.82), and medication (mean 13.08, median 12, min 1, max 67, std 7.07). The combined events per visit have mean 32.81, median 31, min 3, max 119, and std 13.67. The final vocabulary has 62{,}574 nodes (dx: 25{,}711; proc: 36{,}617; med: 246).

\paragraph{eICU.}
Using the same preprocessing protocol, the cohort contains 10{,}766 patients and 25{,}980 visits, with 15{,}214 next-visit prediction pairs for \(T_{\min}=2\). Averaged over 5 seeds, train/val/test contain 8{,}612/1{,}076/1{,}078 patients and 20{,}768/2{,}618/2{,}594 visits. Per-visit event counts are: diagnosis (mean 3.58, median 2, min 1, max 58, std 3.62), procedure (mean 7.62, median 4, min 1, max 148, std 9.42), and medication (mean 18.85, median 17, min 1, max 109, std 10.90). The combined events per visit have mean 30.05, median 25, min 3, max 274, and std 18.36. The final vocabulary has 5{,}912 nodes (dx: 1{,}290; proc: 2{,}638; med: 1{,}984).

\paragraph{Data Splits and Leakage Prevention.} We split the dataset \emph{by patient} into training (80\%), validation (10\%), and test (10\%) sets, ensuring patient-level leakage control (no visits from the same patient across splits). Crucially, all data-driven graph statistics used for temporal stitching, especially lagged PMI, are computed \emph{strictly on the training split} to prevent label leakage. No information from validation or test splits is used in graph construction, model training, or hyperparameter selection.

\paragraph{Split Coverage.}
\label{app:coverage}
We report overlap of unique event nodes and lag-1 transition edges between training and held-out splits (averaged over 5 seeds: 13, 21, 42, 87, 123).  
For \textbf{MIMIC-IV}, validation/test node overlap with training is high (val: 91.97\%, test: 92.17\%), with only 8.03\%/7.83\% unseen nodes, respectively. Transition-edge overlap is lower but still substantial (val: 71.17\%, test: 71.50\%), leaving 28.83\%/28.50\% unseen edges.  
For \textbf{eICU}, node overlap is even higher (val: 97.35\%, test: 97.28\%), with 2.65\%/2.72\% unseen nodes; transition-edge overlap is 80.70\% (val) and 80.64\% (test), with 19.30\%/19.36\% unseen edges.  
Overall, most validation/test nodes and edges are observed during training, while a non-trivial unseen fraction remains, reflecting a realistic distributional shift in longitudinal EHR data. As a result, models must generalize beyond memorized co-occurrence patterns and reason over partially observed transition structure, which is particularly challenging for rare or weakly supported cross-modal dependencies.

\section{Baseline Implementations}
\label{app:baselines}

We compare Risk Horizons against sequential, retrieval-based, hyperbolic, and LLM-based baselines under a unified protocol. Unless otherwise noted, all trainable baselines use embedding dimension $d=64$, patient-level 80/10/10 splits, minimum visits $=2$, and minimum code frequency $=0$.

\paragraph{Sequential Baselines.}
\begin{itemize}
    \item \textbf{Copy Last.} A deterministic persistence baseline that predicts events from the most recent observed visit.

    \item \textbf{Transformer \cite{vaswani2017attention}.} A visit-sequence Transformer over flattened visit event sets, using 2 layers, 4 attention heads, hidden size $64$, and BCE loss.

    \item \textbf{RETAIN \cite{choi2016retain}.} A reverse-time dual-RNN attention model with visit-level and variable-level attention, trained with BCE loss.

    \item \textbf{GRAM \cite{choi2017gram}.} An ontology-aware model that aggregates code and ancestor representations using hierarchy-derived attention.

    \item \textbf{BEHRT \cite{li2020behrt}.} A BERT-style longitudinal EHR encoder with visit-position and modality-type embeddings.

    \item \textbf{Med-BERT \cite{rasmy2021med}.} A medical BERT-style longitudinal encoder over event tokens with visit-position embeddings.

    \item \textbf{HALO \cite{theodorou2023synthesize}.} An autoregressive EHR sequence model adapted to multi-label next-visit prediction.
\end{itemize}

\paragraph{Geometry and Retrieval Baselines.}
\begin{itemize}
    \item \textbf{Euclidean Risk Horizons.} Risk Horizons with the same graph construction and objectives, but replacing hyperbolic embeddings and distances with Euclidean counterparts.

    \item \textbf{Euclidean RAG.} Distance-based Euclidean retrieval without directional cone filtering.

    \item \textbf{+LLM variants.} LLM reranking applied to Euclidean candidate sets using the same constrained scoring protocol as Risk Horizons.
\end{itemize}

\paragraph{Hyperbolic Baselines.}
\begin{itemize}
    \item \textbf{Hyperbolic Linear Probe.} A lightweight multi-label linear classifier trained on frozen Risk Horizons embeddings and visit representations.

    \item \textbf{Hyperbolic NN Global.} A non-parametric nearest-neighbor retrieval baseline using frozen hyperbolic embeddings, without patient-specific Risk Horizon construction.
\end{itemize}

\paragraph{LLM Baselines.}
\begin{itemize}
    \item \textbf{Frequency-candidate reranking.}
    LLMs rerank candidate pools constructed from training-set frequency statistics using the same constrained candidate-scoring protocol as Risk Horizons. Candidate size is set to $K=80$ by default. The evaluated LLM backends include \texttt{meta-llama/Llama-3.1-8B-Instruct} \cite{grattafiori2024llama}, \texttt{meta-llama/Llama-3.3-70B-Instruct} \cite{grattafiori2024llama}, \texttt{epfl-llm/meditron-70b} \cite{chen2023meditron}, and \texttt{mistralai/Mixtral-8x7B-Instruct-v0.1} \cite{jiang2024mixtral}, all accessed through Hugging Face Transformers.
\end{itemize}

\section{Evaluation Metrics}
\label{app:metrics}

We evaluate ranking quality, hierarchy consistency, retrieval quality, and efficiency. Main MIMIC-IV results are reported in Table~\ref{tab:main_multimodal}, extended metrics in Table~\ref{tab:extended_metrics}, and eICU results in Table~\ref{tab:main_eicu}.

\paragraph{Ranking Metrics.}
We report Recall@k and nDCG@k ($k\in\{5,10,20\}$), computed per visit and macro-averaged across modalities. We additionally report MRR and Micro-F1 after threshold tuning on the validation split.

\paragraph{Hierarchy and Retrieval Diagnostics.}
To evaluate structural consistency and candidate-space quality beyond exact matching, we report:
\begin{itemize}
    \item \textbf{Hierarchy Distance (Tree Dist, $\downarrow$):} Average shortest-path distance between predicted and ground-truth events in the hierarchy.
    
    \item \textbf{Ancestor Match@2 ($\uparrow$):} Fraction of predictions sharing an ancestor with the target within two hierarchy levels.
    
    \item \textbf{Candidate Recall ($\uparrow$):} Fraction of ground-truth events covered by the retrieved candidate pool.
    
    \item \textbf{Risk-Horizon Recall@K ($\uparrow$):} Fraction of ground-truth events retained in the top-$K$ retrieved candidates.
    
    \item \textbf{Oracle Recall@K ($\uparrow$):} Upper-bound recall achievable under perfect reranking of the retrieved candidate set.
\end{itemize}

\paragraph{Efficiency.}
We report average evaluation latency per visit step.

\section{Implementation Details}
\label{app:impl}

All models are implemented in PyTorch. Hyperbolic optimization is implemented with \texttt{geoopt}. Experiments are conducted on a single NVIDIA H100 GPU.

\paragraph{General Training Configuration.}
Unless otherwise noted, Risk Horizons uses:
\begin{itemize}
    \item embedding dimension $d=64$,
    \item Poincar\'e ball manifold with initial curvature $c=1.0$,
    \item Riemannian Adam optimizer,
    \item learning rate $10^{-2}$,
    \item batch size $256$,
    \item maximum training epochs $50$,
    \item $10$ negative samples per positive edge,
    \item edge-loss margin $1.0$,
    \item masked-visit loss weight $\alpha=1.0$,
    \item masking ratio $\rho=0.30$,
    \item soft assignment temperature $\beta=5.0$.
\end{itemize}

\paragraph{Graph Construction and Retrieval.}
The main Risk Horizons variant uses:
\begin{itemize}
    \item lag horizon $L=1$,
    \item PMI threshold $\tau=0.0$,
    \item minimum support $\kappa=3$,
    \item bootstrap stability filtering threshold $q=0.5$,
    \item intra-visit and intra-modality stitching enabled,
    \item multi-head Central Events with $H=4$ heads,
    \item multi-cone retrieval with cone aperture $\phi=0.5$.
\end{itemize}

\paragraph{Validation and Model Selection.}
Early stopping is performed using validation Recall@10 with patience $5$. Threshold tuning for Micro-F1 is performed on the validation split prior to final evaluation.

\paragraph{LLM Usage.}
For LLM-augmented variants, the LLM is used strictly as an inference-time reranker over candidates retrieved by Risk Horizons. The LLM cannot introduce new candidates, and no gradients are propagated through the language model.

\begin{figure}[!h]
\centering
\begin{tcolorbox}[
    colback=blue!1,
    colframe=blue!25,
    title=\textbf{Prompt Template for Constrained Candidate Reranking},
    coltitle=black,
    fonttitle=\small,
    boxrule=0.4pt,
    arc=3pt,
    left=7pt, right=7pt, top=6pt, bottom=6pt
]
\small
You are a clinical event prediction assistant.

\vspace{4pt}
\textbf{Task:}\\
Given a patient's longitudinal visit history and a candidate list, select and rank only the events from the provided candidate list that are most likely to occur in the NEXT visit.

\vspace{4pt}
\textbf{Modalities:}
\begin{itemize}[leftmargin=1.2em, itemsep=0pt, topsep=1pt]
    \item dx: diagnosis
    \item proc: procedure
    \item med: medication
\end{itemize}

\textbf{Instructions:}
\begin{itemize}[leftmargin=1.2em, itemsep=0pt, topsep=1pt]
    \item Consider temporal progression (earlier $\rightarrow$ later visits).
    \item Consider cross-modality consistency (diagnoses, procedures, medications).
    \item Favor clinically plausible developments over unrelated or overly generic events.
    \item Select likely events only from the provided candidate list.
    \item Do not introduce, infer, or output any event that is not present in the candidate list.
    \item Keep outputs concise and clinically grounded.
\end{itemize}

\textbf{Output:}\\
Return JSON only in the following format: \\
\texttt{\{"dx": [\{"code": "...", "text": "...", "rank": 1\}], "proc": [...], "med": [...]\}}\\
Use objects of the form \texttt{\{"code": "...", "text": "...", "rank": ...\}} for each selected event.\\
Do not output any other text.

\vspace{4pt}
\textbf{Patient history (oldest $\rightarrow$ newest):}\\
\texttt{\{history\_text\}}

\vspace{4pt}
\textbf{Candidate list:}\\
\texttt{dx: \{dx\_candidates\}}\\
\texttt{proc: \{proc\_candidates\}}\\
\texttt{med: \{med\_candidates\}}

\vspace{4pt}
\textbf{Candidate context:}
\begin{itemize}[leftmargin=1.2em, itemsep=0pt, topsep=1pt]
    \item Candidates are produced by upstream structured retrieval.
    \item The candidate list is intended to constrain prediction to plausible next-visit events.
\end{itemize}

\textbf{Ranked next-visit events (JSON):}
\end{tcolorbox}
\caption{Prompt template used for constrained LLM reranking. The LLM selects and ranks events only from the provided candidate list and is not allowed to expand the candidate set.}
\label{fig:prompt}
\end{figure}

\subsection{LLM Prompting and Scoring Details}
\label{app:llm}

Figure~\ref{fig:prompt} shows the prompt template used for constrained LLM reranking. The LLM receives a linearized patient history together with modality-specific candidate sets retrieved by Risk Horizons and returns ranked predictions restricted to the provided candidates.

We use a listwise reranking formulation in which diagnoses, procedures, and medications are grouped by modality. The prompt instructs the LLM to consider temporal progression and cross-modal consistency while avoiding unsupported candidate expansion. Candidates not selected by the LLM receive the lowest reranking score. LLM outputs are combined with geometric retrieval scores using Eq.~\eqref{eq:rerank_main}.

To control context length and inference cost, we retain only the most recent visit in the linearized patient history. The retrieved candidate pool size is determined by the Risk Horizon retrieval configuration and is typically around 100 candidates per modality in our main experiments. Typical histories contain an average of 177 diagnosis tokens, 46 procedure tokens, and 182 medication tokens. The resulting prompts contain approximately 3{,}500 input tokens on average, with a maximum context budget of 4{,}096 tokens.

Generation is performed with greedy decoding (temperature $T=0$) for deterministic inference. Candidate expansion is explicitly disabled by constraining the LLM to rank only events appearing in the retrieved candidate list. Outputs are restricted to structured JSON responses with a maximum generation budget of 4{,}096 tokens. No in-context examples are provided, and no gradients are propagated through the language model.

All LLM experiments are executed through Hugging Face using identical prompts and decoding configurations across datasets and evaluation splits.

\subsection{Reproducibility}

The anonymized project repository linked in the Introduction contains preprocessing scripts, graph construction pipelines, model implementations, training configurations, and evaluation code required to reproduce the experiments after obtaining dataset access.

All data splits are generated at the patient level using fixed random seeds. To avoid redistributing restricted dataset-derived information, we release the split-generation procedure rather than patient-level split files. All results are averaged over five random seeds, and all data-dependent statistics (e.g., lagged PMI edges) are computed strictly on the training split to prevent information leakage.

\paragraph{Data Accessibility.}
All datasets used in this work are publicly available through PhysioNet\footnote{\url{https://physionet.org/}}. Access to MIMIC-IV and eICU requires credentialed approval and completion of the required data-use training procedures specified by PhysioNet.

\section{Additional Results on MIMIC-IV}
\label{app:mimic}

\subsection{Full Evaluation Results on MIMIC-IV}

Tables~\ref{tab:main_multimodal} and~\ref{tab:extended_metrics} provide the full MIMIC-IV evaluation, including baselines and metrics omitted from the main text for space. These results complement the main analysis by separating three effects: whether a method can recover plausible future events, whether it can rank them well, and whether downstream reranking can compensate for weak candidate construction.

First, the complete table confirms that the advantage of Risk Horizons is not explained by model capacity alone. Sequential baselines remain competitive for frequent and persistent events, especially medications, but their performance varies substantially across modalities. RETAIN, for example, achieves the highest medication nDCG@10, while Med-BERT is strong in overall R@10; however, neither provides consistently strong diagnosis, procedure, and medication prediction. This supports the main-text observation that longitudinal EHR prediction is not only a temporal modeling problem, but also a structured output-space problem.

Second, LLM-only baselines perform poorly when restricted to frequent-event candidate sets. Even large models such as Llama-3.3-70B-Instruct and Meditron-70B remain far below sequential and Risk Horizons variants. This result should be interpreted as a limitation of unconstrained or weakly constrained prediction rather than as a limitation of language models themselves: next-visit prediction requires selecting sparse, multi-label clinical futures from a large and structured vocabulary, and frequent-event candidate sets provide insufficient patient-specific structure.

Third, Euclidean retrieval and Euclidean Risk Horizons remain weak even when paired with LLM reranking. In several cases, adding an LLM to Euclidean candidates does not improve performance and may reduce it. This indicates that downstream reasoning cannot reliably repair a poorly constructed candidate space. By contrast, hyperbolic Risk Horizons without LLM reranking already provides a substantially stronger starting point, suggesting that the main benefit comes from constructing a structured and patient-specific Risk Horizon before final ranking.

\begin{table*}[!h]
\centering
\caption{Multi-modal next-visit prediction on MIMIC-IV (mean $\pm$ std over 5 seeds).}
\label{tab:main_multimodal}
\resizebox{\textwidth}{!}{%
\begin{tabular}{l|cc|cc|cc|cc}
\toprule
& \multicolumn{2}{c|}{\textbf{Dx}} 
& \multicolumn{2}{c|}{\textbf{Proc}} 
& \multicolumn{2}{c|}{\textbf{Med}} 
& \multicolumn{2}{c}{\textbf{Overall}} \\
\textbf{Model} 
& R@10 & nDCG@10 
& R@10 & nDCG@10 
& R@10 & nDCG@10 
& R@10 & nDCG@10 \\
\midrule

\textit{Sequential Baselines} \\
Copy Last
& 0.190 {\small $\pm$ 0.002} & 0.303 {\small $\pm$ 0.001}
& 0.166 {\small $\pm$ 0.030} & 0.152 {\small $\pm$ 0.024}
& 0.452 {\small $\pm$ 0.031} & 0.619 {\small $\pm$ 0.012}
& 0.269 {\small $\pm$ 0.029} & 0.358 {\small $\pm$ 0.012} \\

Transformer (Eucl.)
& 0.141 {\small $\pm$ 0.001} & 0.279 {\small $\pm$ 0.003}
& 0.263 {\small $\pm$ 0.002} & 0.199 {\small $\pm$ 0.011}
& 0.498 {\small $\pm$ 0.010} & 0.687 {\small $\pm$ 0.005}
& 0.301 {\small $\pm$ 0.005} & 0.389 {\small $\pm$ 0.008} \\

BEHRT
& 0.149 {\small $\pm$ 0.007} & 0.293 {\small $\pm$ 0.002}
& 0.280 {\small $\pm$ 0.005} & 0.221 {\small $\pm$ 0.006}
& 0.505 {\small $\pm$ 0.051} & 0.694 {\small $\pm$ 0.010}
& 0.311 {\small $\pm$ 0.038} & 0.402 {\small $\pm$ 0.007} \\

RETAIN
& 0.220 {\small $\pm$ 0.013} & 0.423 {\small $\pm$ 0.019}
& 0.336 {\small $\pm$ 0.016} & 0.268 {\small $\pm$ 0.017}
& 0.532 {\small $\pm$ 0.030} & \cellcolor{best} 0.723 {\small $\pm$ 0.034}
& 0.363 {\small $\pm$ 0.021} & \cellcolor{second} 0.471 {\small $\pm$ 0.023} \\

GRAM
& 0.093 {\small $\pm$ 0.003} & 0.178 {\small $\pm$ 0.003}
& 0.170 {\small $\pm$ 0.008} & 0.119 {\small $\pm$ 0.010}
& 0.479 {\small $\pm$ 0.030} & 0.661 {\small $\pm$ 0.021}
& 0.247 {\small $\pm$ 0.014} & 0.319 {\small $\pm$ 0.011} \\

Med-BERT
& 0.249 {\small $\pm$ 0.014} & 0.303 {\small $\pm$ 0.042}
& \cellcolor{second} 0.380 {\small $\pm$ 0.027} & 0.247 {\small $\pm$ 0.009}
& 0.576 {\small $\pm$ 0.011} & 0.693 {\small $\pm$ 0.012}
& 0.402 {\small $\pm$ 0.017} & 0.414 {\small $\pm$ 0.021} \\

HALO
& 0.128 {\small $\pm$ 0.056} & 0.251 {\small $\pm$ 0.043}
& 0.244 {\small $\pm$ 0.037} & 0.185 {\small $\pm$ 0.060}
& 0.495 {\small $\pm$ 0.074} & 0.683 {\small $\pm$ 0.075}
& 0.289 {\small $\pm$ 0.056} & 0.373 {\small $\pm$ 0.059} \\

\midrule
\textit{LLMs (Top80 Freq. Events)} \\
Llama-3.1-8B-Instruct
& 0.112 {\small $\pm$ 0.001} & 0.135 {\small $\pm$ 0.001}
& 0.096 {\small $\pm$ 0.001} & 0.124 {\small $\pm$ 0.004}
& 0.205 {\small $\pm$ 0.001} & 0.276 {\small $\pm$ 0.002}
& 0.138 {\small $\pm$ 0.001} & 0.178 {\small $\pm$ 0.002} \\

Llama-3.3-70B-Instruct
& 0.196 {\small $\pm$ 0.021} & 0.216 {\small $\pm$ 0.034}
& 0.077 {\small $\pm$ 0.019} & 0.074 {\small $\pm$ 0.012}
& 0.220 {\small $\pm$ 0.009} & 0.382 {\small $\pm$ 0.010}
& 0.164 {\small $\pm$ 0.016} & 0.224 {\small $\pm$ 0.019} \\

Mixtral-8x7B-Instruct-v0.1
& 0.187 {\small $\pm$ 0.039} & 0.148 {\small $\pm$ 0.035}
& 0.068 {\small $\pm$ 0.045} & 0.067 {\small $\pm$ 0.041}
& 0.209 {\small $\pm$ 0.071} & 0.397 {\small $\pm$ 0.060}
& 0.155 {\small $\pm$ 0.052} & 0.204 {\small $\pm$ 0.045} \\

Meditron-70B
& 0.201 {\small $\pm$ 0.030} & 0.205 {\small $\pm$ 0.017}
& 0.129 {\small $\pm$ 0.009} & 0.146 {\small $\pm$ 0.013}
& 0.274 {\small $\pm$ 0.009} & 0.391 {\small $\pm$ 0.004}
& 0.201 {\small $\pm$ 0.016} & 0.247 {\small $\pm$ 0.011} \\

\midrule
\textit{Graph \& Retrieval} \\
Euclidean RAG
& 0.071 {\small $\pm$ 0.006} & 0.142 {\small $\pm$ 0.001}
& 0.127 {\small $\pm$ 0.003} & 0.091 {\small $\pm$ 0.005}
& 0.258 {\small $\pm$ 0.004} & 0.359 {\small $\pm$ 0.002}
& 0.152 {\small $\pm$ 0.005} & 0.197 {\small $\pm$ 0.004} \\

Euc.RAG+Llama-3.1-8B-I.
& 0.063 {\small $\pm$ 0.009} & 0.098 {\small $\pm$ 0.007}
& 0.092 {\small $\pm$ 0.050} & 0.103 {\small $\pm$ 0.009}
& 0.212 {\small $\pm$ 0.093} & 0.239 {\small $\pm$ 0.061}
& 0.122 {\small $\pm$ 0.051} & 0.147 {\small $\pm$ 0.026} \\

Euclidean Risk Horizons
& 0.073 {\small $\pm$ 0.007} & 0.146 {\small $\pm$ 0.009}
& 0.147 {\small $\pm$ 0.002} & 0.097 {\small $\pm$ 0.002}
& 0.244 {\small $\pm$ 0.004} & 0.328 {\small $\pm$ 0.003}
& 0.155 {\small $\pm$ 0.005} & 0.190 {\small $\pm$ 0.005} \\

Euc.RH+Llama-3.1-8B-I.
& 0.081 {\small $\pm$ 0.010} & 0.168 {\small $\pm$ 0.023}
& 0.132 {\small $\pm$ 0.032} & 0.150 {\small $\pm$ 0.039}
& 0.196 {\small $\pm$ 0.031} & 0.296 {\small $\pm$ 0.020}
& 0.136 {\small $\pm$ 0.024} & 0.205 {\small $\pm$ 0.027} \\

\midrule
\textit{Hyperbolic Baselines} \\
Hyperbolic Linear Probe
& 0.129 {\small $\pm$ 0.030} & 0.249 {\small $\pm$ 0.023}
& 0.249 {\small $\pm$ 0.017} & 0.179 {\small $\pm$ 0.030}
& 0.489 {\small $\pm$ 0.034} & 0.678 {\small $\pm$ 0.035}
& 0.289 {\small $\pm$ 0.027} & 0.369 {\small $\pm$ 0.029} \\

Hyperbolic NN Global
& 0.125 {\small $\pm$ 0.063} & 0.243 {\small $\pm$ 0.028}
& 0.166 {\small $\pm$ 0.021} & 0.125 {\small $\pm$ 0.050}
& 0.362 {\small $\pm$ 0.035} & 0.514 {\small $\pm$ 0.025}
& 0.218 {\small $\pm$ 0.040} & 0.294 {\small $\pm$ 0.034} \\

\midrule
\textit{Ours} \\
Risk Horizons (w/o LLM)
& 0.195 {\small $\pm$ 0.005} & 0.392 {\small $\pm$ 0.012}
& 0.298 {\small $\pm$ 0.009} & 0.243 {\small $\pm$ 0.009}
& 0.579 {\small $\pm$ 0.012} & \cellcolor{second} 0.705 {\small $\pm$ 0.014}
& 0.357 {\small $\pm$ 0.013} & 0.447 {\small $\pm$ 0.017} \\

RH+Llama-3.1-8B-I.
& 0.213 {\small $\pm$ 0.050} & \cellcolor{second} 0.453 {\small $\pm$ 0.012}
& 0.298 {\small $\pm$ 0.009} & 0.243 {\small $\pm$ 0.010}
& 0.579 {\small $\pm$ 0.031} & \cellcolor{second} 0.705 {\small $\pm$ 0.047}
& 0.363 {\small $\pm$ 0.030} & 0.467 {\small $\pm$ 0.023} \\

RH+Llama-3.3-70B-I.
& \cellcolor{best} 0.306 {\small $\pm$ 0.032} & \cellcolor{best} 0.496 {\small $\pm$ 0.064}
& 0.368 {\small $\pm$ 0.005} & \cellcolor{second} 0.360 {\small $\pm$ 0.008}
& \cellcolor{best} 0.602 {\small $\pm$ 0.016} & 0.693 {\small $\pm$ 0.014}
& \cellcolor{best} 0.425 {\small $\pm$ 0.018} & \cellcolor{best} 0.516 {\small $\pm$ 0.029} \\

RH+Meditron-70B
& \cellcolor{second} 0.297 {\small $\pm$ 0.011} & 0.417 {\small $\pm$ 0.033}
& \cellcolor{best} 0.389 {\small $\pm$ 0.023} & \cellcolor{best} 0.365 {\small $\pm$ 0.009}
& \cellcolor{second} 0.587 {\small $\pm$ 0.058} & 0.601 {\small $\pm$ 0.027}
& \cellcolor{second} 0.424 {\small $\pm$ 0.031} & 0.461 {\small $\pm$ 0.023} \\
\bottomrule
\end{tabular}%
}
\end{table*}

The extended metrics in Table~\ref{tab:extended_metrics} further support this retrieval--ranking decomposition. Risk Horizons + Llama-3.3-70B-Instruct achieves the strongest Recall@5, Recall@20, Micro-F1, and MRR, indicating that gains persist beyond the top-10 cutoff. The improvement in MRR is particularly informative: geometry-only Risk Horizons already retrieves relevant events effectively, while constrained reranking moves those events earlier in the ranked list. This is consistent with the main-text interpretation that Risk Horizons addresses the candidate-space coverage problem, while LLM reranking primarily improves local ordering within that space.

Finally, the latency results show the expected trade-off between retrieval-only and reranked inference. Geometry-only Risk Horizons is substantially faster, while LLM-based reranking adds inference overhead. This cost is bounded by the compact Risk Horizon, since the LLM scores a restricted candidate set rather than the full vocabulary.

\begin{table*}[!h]
\centering
\caption{Extended evaluation metrics on MIMIC-IV (overall, macro; mean $\pm$ std over 5 seeds).}
\label{tab:extended_metrics}
\resizebox{\textwidth}{!}{%
\begin{tabular}{l|ccccccc}
\toprule
\textbf{Model} 
& R@5 & R@20 
& nDCG@5 & nDCG@20 
& Micro-F1 
& MRR 
& Latency (s/step) \\
\midrule

\textit{Sequential Baselines} \\
Copy Last
& 0.177 {\small $\pm$ 0.013} & 0.337 {\small $\pm$ 0.007}
& 0.361 {\small $\pm$ 0.009} & 0.347 {\small $\pm$ 0.002}
& 0.325 {\small $\pm$ 0.005} & 0.481 {\small $\pm$ 0.002} & 0.001 \\

Transformer (Eucl.) 
& 0.197 {\small $\pm$ 0.004} & 0.424 {\small $\pm$ 0.021}
& 0.408 {\small $\pm$ 0.012} & 0.396 {\small $\pm$ 0.013}
& 0.271 {\small $\pm$ 0.011}
& 0.566 {\small $\pm$ 0.007} & 0.173 \\

BEHRT      
& 0.204 {\small $\pm$ 0.005} & 0.437 {\small $\pm$ 0.003}
& 0.422 {\small $\pm$ 0.007} & 0.410 {\small $\pm$ 0.003}
& 0.278 {\small $\pm$ 0.004}
& 0.583 {\small $\pm$ 0.002} & 0.170 \\

RETAIN              
& 0.240 {\small $\pm$ 0.005} & 0.499 {\small $\pm$ 0.023}
& 0.496 {\small $\pm$ 0.005} & 0.475 {\small $\pm$ 0.012}
& 0.319 {\small $\pm$ 0.003}
& 0.652 {\small $\pm$ 0.005} & 0.176 \\

GRAM              
& 0.152 {\small $\pm$ 0.009} & 0.365 {\small $\pm$ 0.014}
& 0.334 {\small $\pm$ 0.013} & 0.332 {\small $\pm$ 0.018}
& 0.229 {\small $\pm$ 0.006}
& 0.479 {\small $\pm$ 0.007} & 0.172 \\

Med-BERT              
& 0.274 {\small $\pm$ 0.010} & 0.427 {\small $\pm$ 0.050}
& 0.435 {\small $\pm$ 0.031} & 0.431 {\small $\pm$ 0.036}
& 0.268 {\small $\pm$ 0.037}
& 0.583 {\small $\pm$ 0.056} & 0.165 \\

HALO              
& 0.189 {\small $\pm$ 0.045} & 0.409 {\small $\pm$ 0.081}
& 0.391 {\small $\pm$ 0.057} & 0.381 {\small $\pm$ 0.034}
& 0.261 {\small $\pm$ 0.080}
& 0.541 {\small $\pm$ 0.083} & 0.031 \\

\midrule
\textit{LLMs (Top 80 Freq. Events)} \\

Llama-3.1-8B-Instruct 
& 0.076 {\small $\pm$ 0.002} & 0.250 {\small $\pm$ 0.004}
& 0.177 {\small $\pm$ 0.003} & 0.202 {\small $\pm$ 0.003}
& 0.217 {\small $\pm$ 0.002}
& 0.296 {\small $\pm$ 0.003} & 5.211 \\

Llama-3.3-70B-Instruct  
& 0.092 {\small $\pm$ 0.018} & 0.276 {\small $\pm$ 0.021}
& 0.208 {\small $\pm$ 0.025} & 0.232 {\small $\pm$ 0.020}
& 0.224 {\small $\pm$ 0.012}
& 0.332 {\small $\pm$ 0.021} & 5.461 \\

Mixtral-8x7B-Instruct-v0.1 
& 0.090 {\small $\pm$ 0.034} & 0.268 {\small $\pm$ 0.043}
& 0.203 {\small $\pm$ 0.041} & 0.227 {\small $\pm$ 0.038}
& 0.226 {\small $\pm$ 0.030}
& 0.326 {\small $\pm$ 0.039} & 5.667 \\

Meditron-70B  
& 0.116 {\small $\pm$ 0.012} & 0.312 {\small $\pm$ 0.015}
& 0.254 {\small $\pm$ 0.014} & 0.276 {\small $\pm$ 0.013}
& 0.235 {\small $\pm$ 0.009}
& 0.384 {\small $\pm$ 0.014} & 5.893 \\

\midrule
\textit{Graph \& Retrieval} \\

Euclidean RAG        
& 0.086 {\small $\pm$ 0.013} & 0.270 {\small $\pm$ 0.005}
& 0.200 {\small $\pm$ 0.014} & 0.226 {\small $\pm$ 0.003}
& 0.224 {\small $\pm$ 0.011}
& 0.325 {\small $\pm$ 0.003} & 0.140 \\

Euc.RAG+Llama-3.1-8B       
& 0.040 {\small $\pm$ 0.021} & 0.201 {\small $\pm$ 0.035}
& 0.106 {\small $\pm$ 0.028} & 0.135 {\small $\pm$ 0.026}
& 0.196 {\small $\pm$ 0.020}
& 0.216 {\small $\pm$ 0.031} & 6.010 \\

Euclidean Risk Horizons      
& 0.084 {\small $\pm$ 0.001} & 0.277 {\small $\pm$ 0.003}
& 0.187 {\small $\pm$ 0.005} & 0.224 {\small $\pm$ 0.004}
& 0.224 {\small $\pm$ 0.002}
& 0.305 {\small $\pm$ 0.005} & 0.137 \\

Euc.RH+Llama-3.1-8B        
& 0.085 {\small $\pm$ 0.018} & 0.252 {\small $\pm$ 0.027}
& 0.195 {\small $\pm$ 0.026} & 0.221 {\small $\pm$ 0.024}
& 0.230 {\small $\pm$ 0.018}
& 0.316 {\small $\pm$ 0.026} & 5.987 \\

\midrule
\textit{Hyperbolic Baselines} \\

Hyperbolic Linear
& 0.197 {\small $\pm$ 0.017} & 0.367 {\small $\pm$ 0.016}
& 0.352 {\small $\pm$ 0.010} & 0.445 {\small $\pm$ 0.014}
& 0.284 {\small $\pm$ 0.012}
& 0.537 {\small $\pm$ 0.013}
& 0.077 \\

Hyperbolic NN Global
& 0.145 {\small $\pm$ 0.021} & 0.303 {\small $\pm$ 0.020}
& 0.265 {\small $\pm$ 0.018} & 0.358 {\small $\pm$ 0.019}
& 0.232 {\small $\pm$ 0.016}
& 0.501 {\small $\pm$ 0.014}
& 0.073 \\

\midrule
\textit{Ours} \\

RH (w/o LLM backend)             
& 0.223 {\small $\pm$ 0.009} & 0.460 {\small $\pm$ 0.003}
& 0.465 {\small $\pm$ 0.007} & 0.442 {\small $\pm$ 0.006}
& 0.320 {\small $\pm$ 0.006}
& 0.613 {\small $\pm$ 0.010} 
& 0.319 \\

RH+Llama-3.1-8B-I.       
& 0.228 {\small $\pm$ 0.030} 
& 0.465 {\small $\pm$ 0.037}
& 0.454 {\small $\pm$ 0.022} 
& 0.431 {\small $\pm$ 0.025}
& 0.307 {\small $\pm$ 0.019}
& 0.622 {\small $\pm$ 0.025} 
& 5.967 \\

RH+Llama-3.3-70B-I.       
& \cellcolor{best} \textbf{0.279} {\small $\pm$ 0.011} 
& \cellcolor{best} \textbf{0.566} {\small $\pm$ 0.018}
& \cellcolor{second} 0.519 {\small $\pm$ 0.034} 
& \cellcolor{second} 0.503 {\small $\pm$ 0.029}
& \cellcolor{best} \textbf{0.338} {\small $\pm$ 0.012}
& \cellcolor{best} \textbf{0.704} {\small $\pm$ 0.028} 
& 6.781 \\

RH+Meditron-70B                    
& \cellcolor{second} 0.274 {\small $\pm$ 0.023} 
& \cellcolor{second} 0.558 {\small $\pm$ 0.032}
& \cellcolor{best} \textbf{0.532} {\small $\pm$ 0.031} 
& \cellcolor{best} \textbf{0.512} {\small $\pm$ 0.027}
& \cellcolor{second} 0.327 {\small $\pm$ 0.020}
& \cellcolor{second} 0.686 {\small $\pm$ 0.031} 
& 6.923 \\

\bottomrule
\end{tabular}%
}
\end{table*}

\subsection{Additional Ablation and Sensitivity Studies}

Table~\ref{tab:ablation_full} analyzes the contribution of the main components of Risk Horizons. All components contribute substantially to performance, supporting the central hypothesis that accurate next-visit prediction depends primarily on constructing structured and temporally coherent candidate spaces before final ranking.

\textbf{Cross-modal temporal stitching is essential.}
Removing lagged PMI stitching causes a near-complete collapse in performance, reducing overall R@10 from 0.357 to 0.005 and nDCG@10 from 0.447 to 0.004. This indicates that deterministic coding hierarchies alone are insufficient for constructing clinically meaningful future hypothesis spaces. Most predictive signal instead arises from inferred cross-modal and temporal dependencies between diagnoses, procedures, and medications.

\textbf{Directional retrieval improves hypothesis-space quality.}
Replacing cone-constrained retrieval with distance-only retrieval reduces overall R@10 from 0.357 to 0.180 and nDCG@10 from 0.447 to 0.206. This suggests that directional constraints are important for modeling asymmetric clinical progression. Without directional retrieval, candidate spaces remain semantically related but become substantially less predictive of future transitions.

\textbf{Masked-visit denoising stabilizes latent clinical states.}
Removing masked-visit reconstruction reduces overall R@10 from 0.357 to 0.115 and nDCG@10 from 0.447 to 0.124. This indicates that denoising-based Central Event training improves robustness under sparse and noisy multi-modal trajectories by encouraging visit representations to preserve predictable clinical structure rather than memorizing individual observations.

\begin{table}[!h]
\centering
\caption{Central Event Weighting Strategies.}
\label{tab:central_weights}
\scalebox{0.8}{
\begin{tabular}{l|cc}
\toprule
\textbf{Method} & R@10 & nDCG@10 \\
\midrule
\textit{Central Events Weighting} & & \\
idf weighting & 0.194 & 0.224 \\
attention weighting & 0.189 & 0.217 \\
modality weighting & 0.192 & 0.220 \\
\bottomrule
\end{tabular}
}
\end{table}

\paragraph{Central-event weighting.}
The central-event weighting results further suggest that more complex weighting schemes do not automatically improve the latent visit representation. IDF, attention-based, and modality-level weighting all underperform uniform weighting. One possible explanation is that the downstream Risk Horizon already imposes strong structural constraints, so additional weighting can overemphasize rare or locally salient codes at the expense of stable patient-state representation.

\paragraph{Sensitivity Analysis.} The sensitivity analysis shows that Risk Horizons is robust to moderate hyperparameter variation but sensitive to choices that damage candidate-space quality. Increasing the number of Central Event heads improves performance up to a moderate value and then saturates, suggesting that multiple heads help separate heterogeneous clinical factors but excessive capacity provides limited benefit. The PMI threshold has a relatively mild effect, while aggressive stability filtering substantially degrades performance, likely because useful but weakly observed temporal associations are removed. Cone aperture exhibits a similar trade-off: moderate apertures preserve directional selectivity, whereas overly broad cones admit semantically related but temporally less plausible candidates. Overall, the sensitivity results reinforce the main conclusion that performance depends on constructing candidate spaces that are broad enough to cover plausible futures but constrained enough to remain clinically coherent.

\begin{figure*}[!h]
\vspace{-0.1cm}
\centering
\includegraphics[width=1\linewidth]{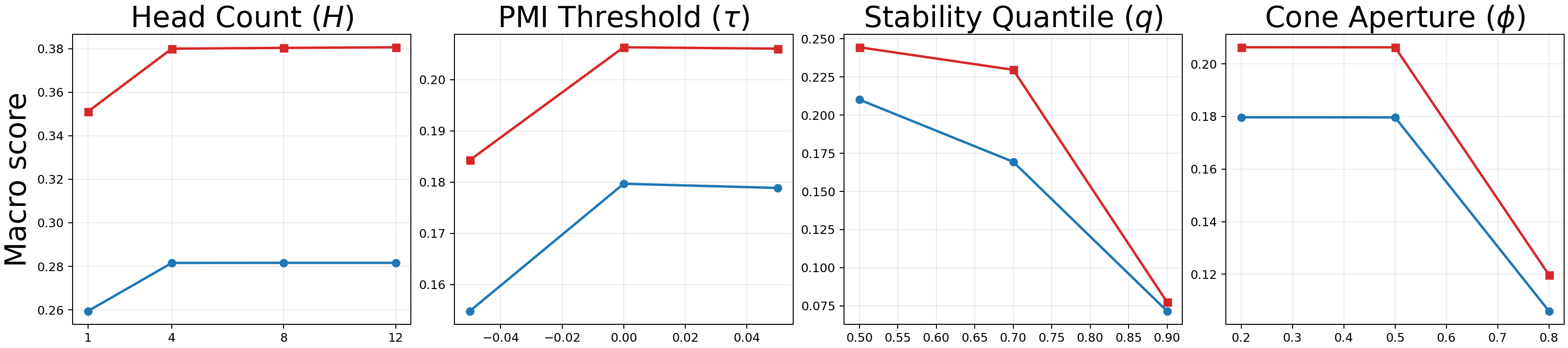}
\vspace{-0.3cm}
\caption{Sensitivity analysis: Recall@10 \textcolor{blue}{(blue)} and nDCG@10 \textcolor{red}{(red)}. 
From left to right: (a) number of heads $H$, 
(b) PMI Threshold $\tau$,
(c) Stability $q$,
(d) cone aperture $\phi$.} 
\label{fig:sensitivity}
\vspace{-0.3cm}
\end{figure*}

\subsection{Repeated vs.\ Novel Event Analysis}
\label{app:repeated_novel}

To better understand the behavior of Risk Horizons, we analyze prediction performance separately for repeated events (events already observed in the patient history) and novel events (events not previously observed in the history). This distinction is particularly important in longitudinal EHR prediction because many future events correspond either to persistent chronic conditions or to clinically coherent progression patterns, while others arise from weakly observed, exogenous, or highly sparse transitions.

Table~\ref{tab:repeated_novel} reports Recall separately for repeated and novel events on MIMIC-IV using the geometry-only Risk Horizons variant. Across all modalities, performance is substantially higher for repeated events than for novel events. This gap is especially pronounced for diagnoses, where repeated-event Recall reaches 0.646 while novel-event Recall is only 0.084.

\begin{table}[!h]
\centering
\caption{Recall on repeated vs.\ novel events on MIMIC-IV. Repeated events are events already present in the patient history; novel events are newly introduced events not previously observed in the trajectory.}
\label{tab:repeated_novel}
\resizebox{0.8\linewidth}{!}{%
\begin{tabular}{l|cc|cc|c}
\toprule
& \multicolumn{2}{c|}{\textbf{Event Distribution}}
& \multicolumn{2}{c|}{\textbf{Recall}}
& \textbf{Weakly Supported} \\
\textbf{Modality}
& Repeated Share
& Novel Share
& Repeated
& Novel
& Novel Events \\
\midrule
Dx
& 28.2\%
& 71.8\%
& 0.646
& 0.084
& 2.2\% \\

Proc
& 13.4\%
& 86.6\%
& 0.707
& 0.171
& 6.9\% \\

Med
& 66.1\%
& 33.9\%
& 0.854
& 0.339
& 0.0\% \\
\bottomrule
\end{tabular}%
}
\end{table}

These findings are consistent with the structured retrieval formulation of Risk Horizons. The framework is most effective when future events follow coherent longitudinal progression and can be recovered through structured candidate construction. In such settings, the retrieved Risk Horizon successfully captures persistent conditions, evolutionarily related diagnoses, and temporally associated medications and procedures.

In contrast, novel-event prediction remains substantially harder, particularly for diagnoses and procedures. Many novel events correspond to traumatic injuries, acute complications, unexpected interventions, or weakly observed transitions with limited historical support. Such events are intrinsically difficult to anticipate from prior longitudinal structure alone. Importantly, even when exact novel events are missed, predictions often remain clinically and hierarchically related to the underlying patient state, consistent with the qualitative examples discussed in Section~\ref{app:qualitative}.

The final column of Table~\ref{tab:repeated_novel} reports the fraction of novel events that are weakly supported in the analyzed trajectories, meaning that they never appear in the historical portion of any patient trajectory within the analyzed samples. Although these events account for only a minority of novel targets (2.2\% for diagnoses and 6.9\% for procedures), they are almost never recovered by the model. Many correspond to rare acute conditions, traumatic injuries, or unexpected interventions with limited recoverable longitudinal structure.

Overall, these findings suggest that the primary strength of Risk Horizons lies in recovering clinically plausible progression trajectories through structured retrieval, while highly novel or exogenous transitions remain a fundamental challenge for longitudinal EHR forecasting more broadly.

\subsection{Geometry of the Learned Retrieval Space}

We analyze the geometry induced by Risk Horizons to better understand how hierarchical structure and cross-modal temporal associations shape the retrieved hypothesis space.

\begin{figure}[!h]
    \centering
    \includegraphics[width=1\linewidth]{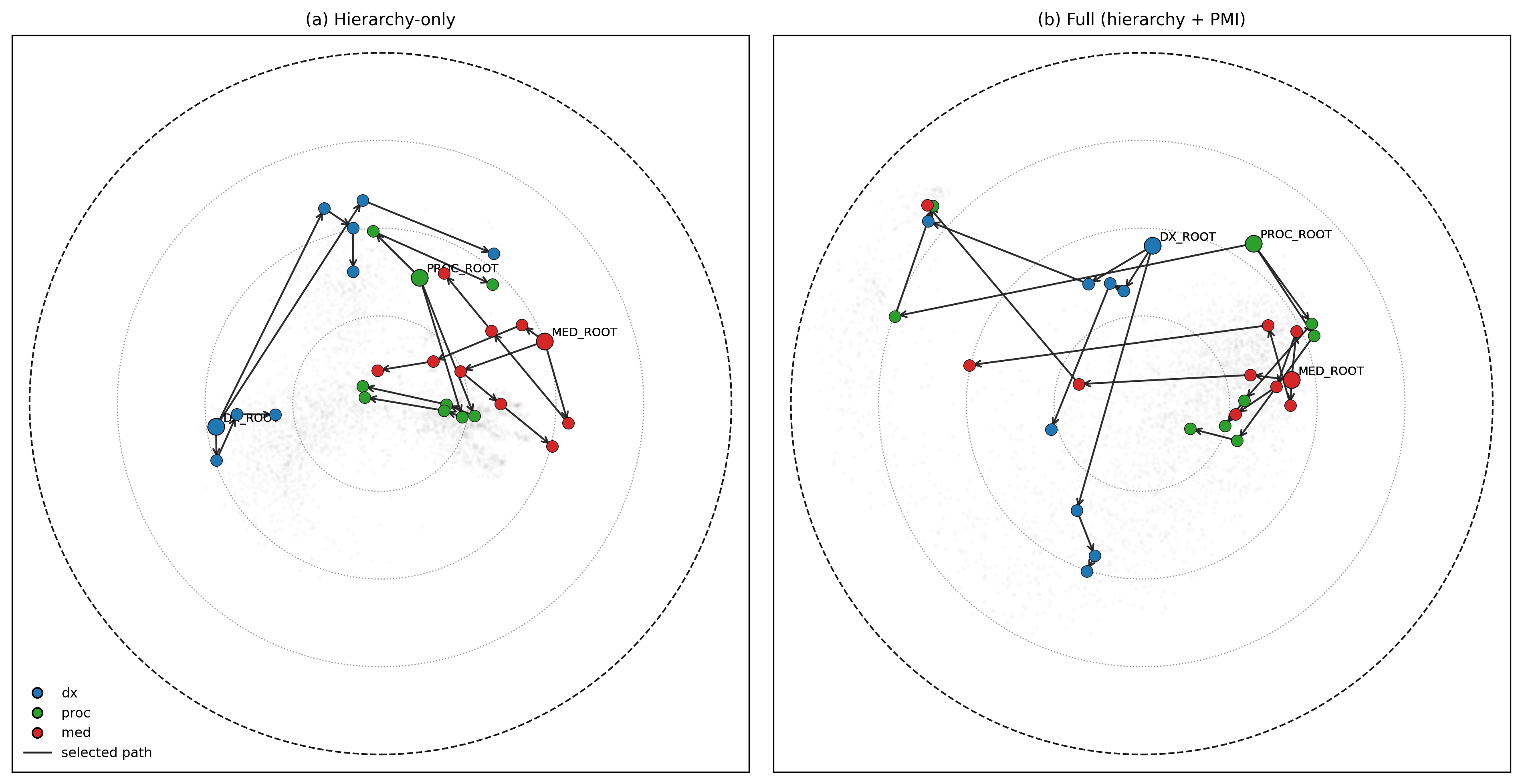}
    \caption{Hierarchy-only vs.\ full embedding (with PMI edges).}
    \label{fig:compare_disk}
\end{figure}

\paragraph{Global hierarchy and local retrieval structure.}
Figure~\ref{fig:compare_disk} compares embeddings learned using hierarchy edges alone versus the full graph with cross-modal PMI stitching. The hierarchy-only embedding forms a predominantly tree-like radial organization characteristic of hyperbolic representations. After introducing PMI edges, the embedding develops localized multimodal neighborhoods while still preserving large-scale hierarchical structure. Rather than collapsing the hierarchy, cross-modal stitching selectively deforms local regions of the space, bringing temporally associated diagnoses, procedures, and medications into closer proximity. This supports the central retrieval objective of Risk Horizons: preserving global semantic organization while constructing clinically coherent local candidate regions.

\begin{figure}[!h]
    \centering
    \includegraphics[width=0.52\linewidth]{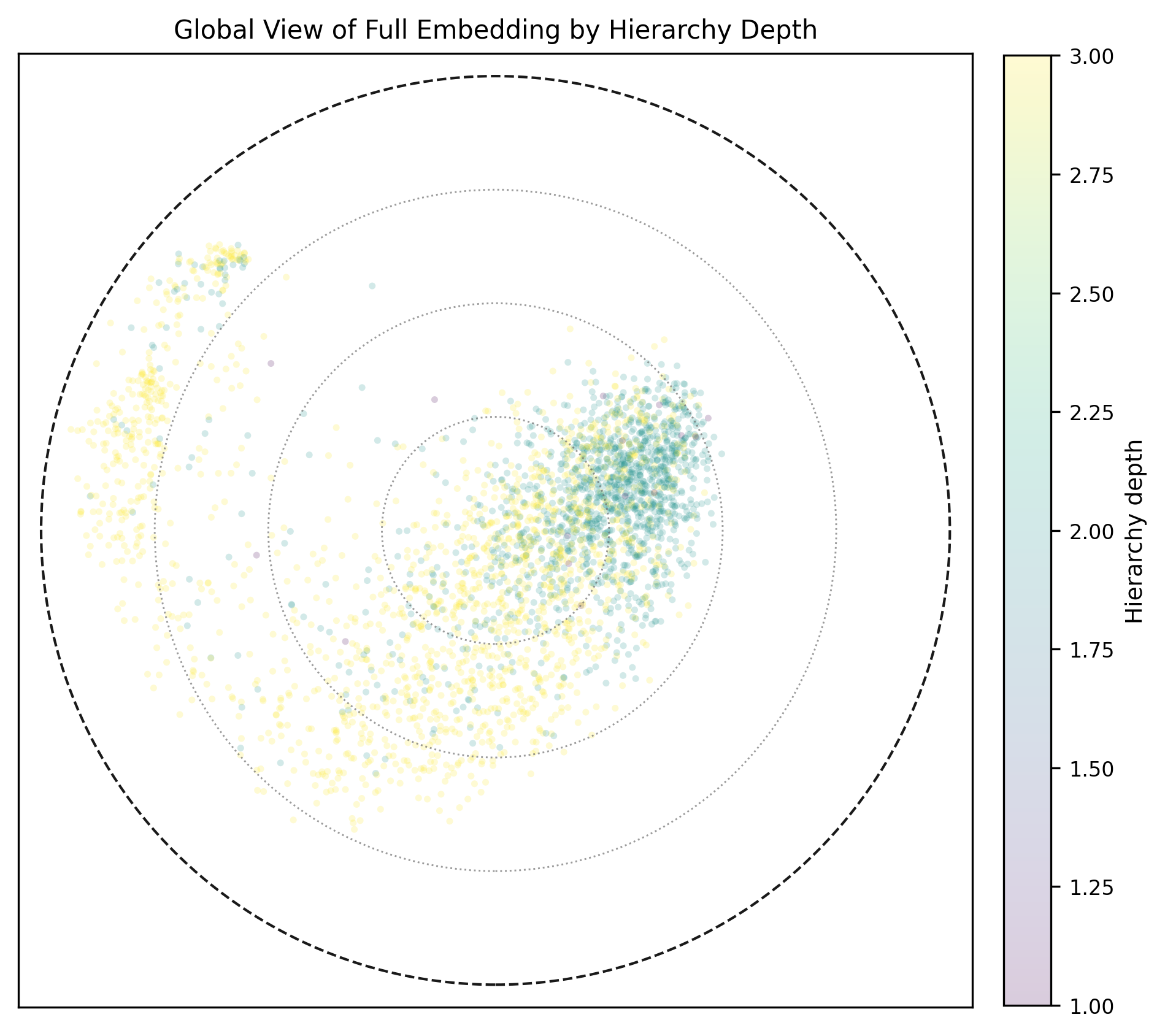}
    \includegraphics[width=0.46\linewidth]{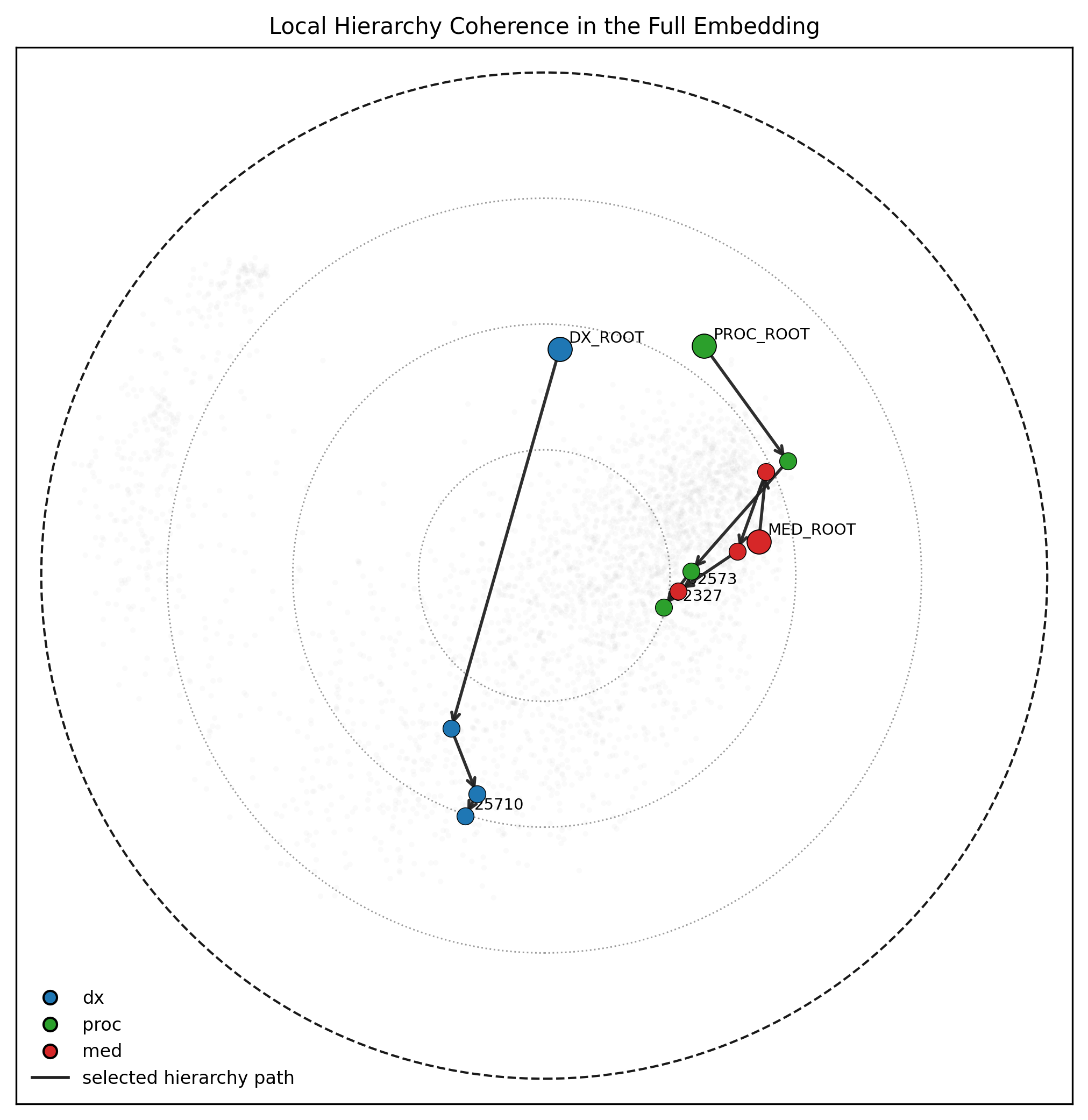}
    \caption{Left: Full embedding colored by hierarchy depth. Right: Selected hierarchical paths in the full embedding.}
    \label{fig:depth_disk}
\end{figure}

Figure~\ref{fig:depth_disk} (left) further shows that hierarchy depth remains approximately organized along the radial direction of the Poincar\'e disk, with deeper concepts generally positioned closer to the boundary. Although PMI stitching introduces local deviations from strict tree geometry, the embedding retains substantial hierarchical organization even after multimodal integration.

\begin{figure}[!h]
    \centering
    \includegraphics[width=0.8\linewidth]{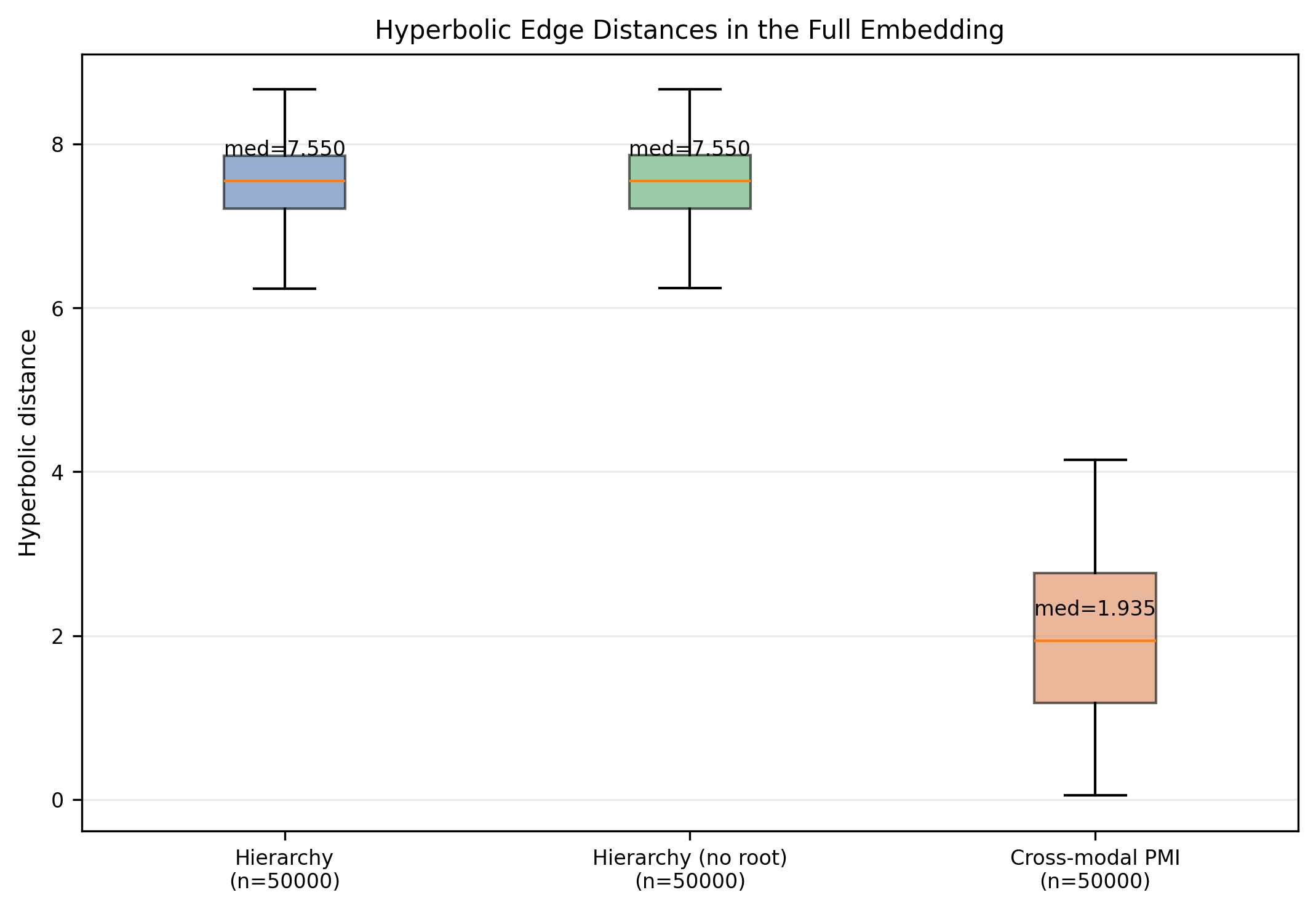}
    \caption{Hyperbolic distance distributions for hierarchy and PMI edges.}
    \label{fig:edge_dist}
\end{figure}

\paragraph{Edge-type separation and local hierarchy preservation.}
Figure~\ref{fig:edge_dist} shows a strong geometric separation between hierarchy and PMI edges. Hierarchical relations occupy substantially larger hyperbolic distances (median $\approx 7.55$), while PMI edges remain highly localized (median $\approx 1.93$). This indicates that cross-modal temporal associations primarily act as local retrieval shortcuts rather than globally distorting the hierarchy.

Figure~\ref{fig:depth_disk} (right) visualizes representative root-to-leaf paths in the full embedding. Despite local deformation induced by cross-modal stitching, hierarchical trajectories remain approximately radial and locally coherent. Together, these observations suggest that Risk Horizons preserves enough global hierarchy for semantic organization while introducing sufficient local flexibility to support clinically meaningful retrieval neighborhoods.

\section{Evaluation Results on eICU}
\label{app:eicu}

\begin{table*}[!ht]
\centering
\caption{Multi-modal next-visit prediction on eICU (mean $\pm$ std over 5 seeds).}
\label{tab:main_eicu}
\resizebox{\textwidth}{!}{%
\begin{tabular}{l|cc|cc|cc|cc}
\toprule
& \multicolumn{2}{c|}{\textbf{Dx}} 
& \multicolumn{2}{c|}{\textbf{Proc}} 
& \multicolumn{2}{c|}{\textbf{Med}} 
& \multicolumn{2}{c}{\textbf{Overall}} \\
\textbf{Model} 
& R@10 & nDCG@10 
& R@10 & nDCG@10 
& R@10 & nDCG@10 
& R@10 & nDCG@10 \\
\midrule

\textit{Sequential Baselines} \\

Copy Last
& 0.273 {\small $\pm$ 0.002} & 0.263 {\small $\pm$ 0.003}
& 0.170 {\small $\pm$ 0.018} & 0.175 {\small $\pm$ 0.014}
& 0.249 {\small $\pm$ 0.021} & 0.421 {\small $\pm$ 0.011}
& 0.231 {\small $\pm$ 0.012} & 0.286 {\small $\pm$ 0.009} \\

Transformer (Eucl.)
& 0.381 {\small $\pm$ 0.006} & 0.300 {\small $\pm$ 0.004}
& 0.251 {\small $\pm$ 0.005} & 0.241 {\small $\pm$ 0.009}
& 0.240 {\small $\pm$ 0.010} & 0.458 {\small $\pm$ 0.006}
& 0.291 {\small $\pm$ 0.007} & 0.333 {\small $\pm$ 0.006} \\

BEHRT
& 0.386 {\small $\pm$ 0.007} & 0.302 {\small $\pm$ 0.006}
& 0.257 {\small $\pm$ 0.006} & 0.241 {\small $\pm$ 0.008}
& 0.241 {\small $\pm$ 0.009} & 0.462 {\small $\pm$ 0.007}
& 0.295 {\small $\pm$ 0.007} & 0.335 {\small $\pm$ 0.007} \\

RETAIN
& 0.505 {\small $\pm$ 0.011} & 0.412 {\small $\pm$ 0.012}
& 0.346 {\small $\pm$ 0.010} & 0.326 {\small $\pm$ 0.009}
& \cellcolor{second} 0.292 {\small $\pm$ 0.013} & \cellcolor{best} 0.533 {\small $\pm$ 0.010}
& 0.381 {\small $\pm$ 0.019} & \cellcolor{second} 0.424 {\small $\pm$ 0.010} \\

GRAM
& 0.335 {\small $\pm$ 0.008} & 0.241 {\small $\pm$ 0.007}
& 0.210 {\small $\pm$ 0.009} & 0.194 {\small $\pm$ 0.008}
& 0.220 {\small $\pm$ 0.012} & 0.441 {\small $\pm$ 0.009}
& 0.255 {\small $\pm$ 0.008} & 0.292 {\small $\pm$ 0.008} \\

Med-BERT
& 0.426 {\small $\pm$ 0.009} & 0.337 {\small $\pm$ 0.010}
& 0.288 {\small $\pm$ 0.008} & 0.272 {\small $\pm$ 0.009}
& 0.252 {\small $\pm$ 0.011} & 0.480 {\small $\pm$ 0.010}
& 0.322 {\small $\pm$ 0.009} & 0.363 {\small $\pm$ 0.009} \\

HALO
& 0.443 {\small $\pm$ 0.013} & 0.353 {\small $\pm$ 0.014}
& 0.310 {\small $\pm$ 0.011} & 0.282 {\small $\pm$ 0.012}
& 0.261 {\small $\pm$ 0.014} & 0.485 {\small $\pm$ 0.012}
& 0.338 {\small $\pm$ 0.012} & 0.373 {\small $\pm$ 0.013} \\

\midrule
\textit{LLMs (Top 80 Freq. Events)} \\

Llama-3.1-8B-Instruct
& 0.195 {\small $\pm$ 0.041} & 0.289 {\small $\pm$ 0.043}
& 0.169 {\small $\pm$ 0.032} & 0.173 {\small $\pm$ 0.030}
& 0.241 {\small $\pm$ 0.012} & 0.310 {\small $\pm$ 0.024}
& 0.202 {\small $\pm$ 0.029} & 0.257 {\small $\pm$ 0.032} \\

Llama-3.3-70B-Instruct
& 0.291 {\small $\pm$ 0.035} & 0.311 {\small $\pm$ 0.051}
& 0.247 {\small $\pm$ 0.034} & 0.211 {\small $\pm$ 0.050}
& 0.239 {\small $\pm$ 0.033} & 0.479 {\small $\pm$ 0.039}
& 0.259 {\small $\pm$ 0.034} & 0.334 {\small $\pm$ 0.044} \\

Mixtral-8x7B-Instruct-v0.1
& 0.283 {\small $\pm$ 0.027} & 0.297 {\small $\pm$ 0.032}
& 0.231 {\small $\pm$ 0.025} & 0.152 {\small $\pm$ 0.021}
& 0.199 {\small $\pm$ 0.030} & 0.269 {\small $\pm$ 0.026}
& 0.238 {\small $\pm$ 0.026} & 0.239 {\small $\pm$ 0.025} \\

Meditron-70B
& 0.297 {\small $\pm$ 0.032} & 0.312 {\small $\pm$ 0.039}
& 0.249 {\small $\pm$ 0.039} & 0.198 {\small $\pm$ 0.024}
& 0.247 {\small $\pm$ 0.033} & 0.451 {\small $\pm$ 0.020}
& 0.264 {\small $\pm$ 0.033} & 0.320 {\small $\pm$ 0.027} \\

\midrule
\textit{Graph \& Retrieval} \\

Euclidean RAG
& 0.300 {\small $\pm$ 0.009} & 0.220 {\small $\pm$ 0.007}
& 0.163 {\small $\pm$ 0.011} & 0.163 {\small $\pm$ 0.009}
& 0.125 {\small $\pm$ 0.014} & 0.222 {\small $\pm$ 0.010}
& 0.196 {\small $\pm$ 0.010} & 0.202 {\small $\pm$ 0.008} \\

Euc.RAG+Llama-3.1-8B
& 0.297 {\small $\pm$ 0.019} & 0.219 {\small $\pm$ 0.017}
& 0.176 {\small $\pm$ 0.021} & 0.173 {\small $\pm$ 0.018}
& 0.139 {\small $\pm$ 0.024} & 0.262 {\small $\pm$ 0.020}
& 0.204 {\small $\pm$ 0.020} & 0.218 {\small $\pm$ 0.018} \\

Euclidean Risk Horizons
& 0.313 {\small $\pm$ 0.006} & 0.240 {\small $\pm$ 0.009}
& 0.183 {\small $\pm$ 0.006} & 0.182 {\small $\pm$ 0.010}
& 0.144 {\small $\pm$ 0.013} & 0.258 {\small $\pm$ 0.021}
& 0.213 {\small $\pm$ 0.008} & 0.227 {\small $\pm$ 0.013} \\

Euc.RH+Llama-3.1-8B
& 0.187 {\small $\pm$ 0.027} & 0.246 {\small $\pm$ 0.014}
& 0.154 {\small $\pm$ 0.030} & 0.201 {\small $\pm$ 0.024}
& 0.258 {\small $\pm$ 0.022} & 0.274 {\small $\pm$ 0.019}
& 0.200 {\small $\pm$ 0.025} & 0.240 {\small $\pm$ 0.019} \\

\midrule
\textit{Hyperbolic Baselines} \\

Hyperbolic Linear Probe
& 0.460 {\small $\pm$ 0.017} & 0.367 {\small $\pm$ 0.011}
& 0.307 {\small $\pm$ 0.009} & 0.291 {\small $\pm$ 0.010}
& 0.259 {\small $\pm$ 0.012} & 0.484 {\small $\pm$ 0.014}
& 0.342 {\small $\pm$ 0.012} & 0.381 {\small $\pm$ 0.008} \\

Hyperbolic NN Global
& 0.407 {\small $\pm$ 0.008} & 0.312 {\small $\pm$ 0.009}
& 0.241 {\small $\pm$ 0.007} & 0.241 {\small $\pm$ 0.018}
& 0.249 {\small $\pm$ 0.011} & 0.429 {\small $\pm$ 0.012}
& 0.299 {\small $\pm$ 0.008} & 0.327 {\small $\pm$ 0.011} \\

\midrule
\textit{Ours} \\

Risk Horizons (w/o LLM)
& 0.462 {\small $\pm$ 0.007} & 0.378 {\small $\pm$ 0.007}
& 0.282 {\small $\pm$ 0.014} & 0.271 {\small $\pm$ 0.006}
& 0.274 {\small $\pm$ 0.005} & 0.483 {\small $\pm$ 0.009}
& 0.339 {\small $\pm$ 0.015} & 0.377 {\small $\pm$ 0.004} \\

RH+Llama-3.1-8B-I.
& 0.469 {\small $\pm$ 0.049} & 0.363 {\small $\pm$ 0.023}
& 0.285 {\small $\pm$ 0.022} & 0.287 {\small $\pm$ 0.031}
& 0.276 {\small $\pm$ 0.036} & 0.489 {\small $\pm$ 0.043}
& 0.343 {\small $\pm$ 0.030} & 0.380 {\small $\pm$ 0.034} \\

RH+Llama-3.3-70B-I.
& \cellcolor{best} 0.519 {\small $\pm$ 0.029} & \cellcolor{best} 0.463 {\small $\pm$ 0.019}
& \cellcolor{second} 0.388 {\small $\pm$ 0.022} & \cellcolor{best} 0.333 {\small $\pm$ 0.045}
& \cellcolor{best} 0.296 {\small $\pm$ 0.031} & \cellcolor{second} 0.502 {\small $\pm$ 0.039}
& \cellcolor{best} 0.401 {\small $\pm$ 0.027} & \cellcolor{best} 0.433 {\small $\pm$ 0.030} \\

RH+Meditron-70B
& \cellcolor{second} 0.506 {\small $\pm$ 0.044} & \cellcolor{second} 0.419 {\small $\pm$ 0.043}
& \cellcolor{best} 0.396 {\small $\pm$ 0.020} & \cellcolor{second} 0.327 {\small $\pm$ 0.034}
& \cellcolor{second} 0.292 {\small $\pm$ 0.036} & 0.497 {\small $\pm$ 0.043}
& \cellcolor{second} 0.398 {\small $\pm$ 0.033} & 0.414 {\small $\pm$ 0.039} \\

\bottomrule
\end{tabular}%
}
\end{table*}

Table~\ref{tab:main_eicu} reports multi-modal next-visit prediction results on eICU. Overall, the trends remain consistent with MIMIC-IV: structure-aware retrieval substantially improves next-visit prediction, while constrained reranking provides additional gains when the retrieved candidate space is sufficiently coherent.

Compared to MIMIC-IV, however, the relative gap between geometry-only and reranked variants is smaller. Geometry-only Risk Horizons already performs strongly on diagnoses (0.462 R@10, 0.378 nDCG@10), outperforming all sequential baselines and both hyperbolic baselines. Adding constrained reranking still improves overall performance, reaching 0.401 overall R@10 and 0.433 overall nDCG@10 with Llama-3.3-70B-Instruct, but the gains are noticeably smaller than those observed on MIMIC-IV. This suggests that, on eICU, candidate retrieval remains the dominant factor, while downstream reranking becomes comparatively less important.

The behavior of the Euclidean and hyperbolic baselines further supports this interpretation. Euclidean retrieval methods remain substantially weaker, even when paired with LLM reranking, indicating that downstream reasoning alone cannot compensate for poorly structured candidate spaces. At the same time, the gap between Hyperbolic NN Global and full Risk Horizons is narrower than on MIMIC-IV, suggesting that eICU trajectories contain weaker hierarchical and cross-modal structure overall. In this setting, much of the predictive signal can already be recovered from local geometric proximity, reducing the relative advantage of more sophisticated horizon construction.

The sequential baselines also behave differently on eICU. RETAIN performs particularly strongly overall, while Copy Last remains competitive for medications, reflecting stronger short-term event persistence and lower effective vocabulary complexity. LLM-only baselines also improve relative to their MIMIC-IV performance, likely because the eICU prediction space is substantially smaller and less diverse. Nevertheless, they still remain well below structured retrieval methods, reinforcing that unconstrained prediction remains difficult even in reduced-vocabulary settings.

These differences are consistent with the underlying dataset characteristics. Compared to MIMIC-IV, eICU contains a substantially smaller and less standardized vocabulary, including synthetic procedure identifiers and non-ATC medication coding. Diagnosis trajectories are also sparser and more heterogeneous across institutions. As a result, the latent hierarchy is weaker, temporal transitions are noisier, and clinically coherent future states are more difficult to organize into sharply separated Risk Horizons. Under these conditions, geometry-aware retrieval remains beneficial, but the additional gains from constrained reranking become smaller because the retrieved candidate space itself is less structured.

Table~\ref{tab:hierarchy_metrics_eicu} further supports this interpretation. Risk Horizons consistently improves hierarchy distance, candidate recall, and Risk-Horizon Recall across all modalities relative to Hyperbolic NN Global, indicating that the retrieved horizons remain more structurally coherent even in the noisier eICU setting. However, the absolute gaps are smaller than on MIMIC-IV, especially for Ancestor Match and Oracle@10, again suggesting that eICU contains weaker recoverable hierarchy structure overall. Importantly, Oracle@10 remains substantially higher than the achieved Recall@10 across modalities, indicating that ranking within the retrieved horizon remains a partially unsolved problem even when candidate retrieval is strong.

\begin{table*}[!h]
\centering
\caption{Hierarchy consistency and candidate-space quality on eICU (mean $\pm$ std over 5 seeds).}
\label{tab:hierarchy_metrics_eicu}
\resizebox{\textwidth}{!}{%
\begin{tabular}{ll|cc|cccc}
\toprule
\textbf{Model} 
& \textbf{Type}
& Hier. Dist $\downarrow$ 
& Anc. Match $\uparrow$
& Cand. Rec. $\uparrow$
& RH@10 $\uparrow$
& RH@40 $\uparrow$
& Oracle@10 $\uparrow$ \\
\midrule

\multirow{3}{*}{Hyper. NN Global}
& Dx
& 1.337 {\small $\pm$ 0.014}
& 0.978 {\small $\pm$ 0.003}
& 0.870 {\small $\pm$ 0.008}
& 0.407 {\small $\pm$ 0.011}
& 0.707 {\small $\pm$ 0.010}
& 0.863 {\small $\pm$ 0.006} \\
& Proc
& 1.537 {\small $\pm$ 0.019}
& 0.956 {\small $\pm$ 0.005}
& 0.660 {\small $\pm$ 0.013}
& 0.241 {\small $\pm$ 0.010}
& 0.458 {\small $\pm$ 0.014}
& 0.641 {\small $\pm$ 0.012} \\
& Med
& 1.902 {\small $\pm$ 0.027}
& 0.986 {\small $\pm$ 0.002}
& 0.740 {\small $\pm$ 0.011}
& 0.249 {\small $\pm$ 0.009}
& 0.519 {\small $\pm$ 0.013}
& 0.586 {\small $\pm$ 0.010} \\

\midrule

\multirow{3}{*}{Risk Horizons}
& Dx
& 1.295 {\small $\pm$ 0.012}
& 0.980 {\small $\pm$ 0.002}
& 0.887 {\small $\pm$ 0.007}
& 0.452 {\small $\pm$ 0.010}
& 0.724 {\small $\pm$ 0.009}
& 0.890 {\small $\pm$ 0.005} \\
& Proc
& 1.493 {\small $\pm$ 0.017}
& 0.959 {\small $\pm$ 0.004}
& 0.694 {\small $\pm$ 0.012}
& 0.269 {\small $\pm$ 0.011}
& 0.486 {\small $\pm$ 0.013}
& 0.667 {\small $\pm$ 0.011} \\
& Med
& 1.828 {\small $\pm$ 0.024}
& 0.984 {\small $\pm$ 0.003}
& 0.766 {\small $\pm$ 0.010}
& 0.271 {\small $\pm$ 0.008}
& 0.540 {\small $\pm$ 0.012}
& 0.599 {\small $\pm$ 0.009} \\

\bottomrule
\end{tabular}%
}
\end{table*}

\subsection{Embedding Geometry on eICU}

We visualize the learned hyperbolic embeddings on eICU to examine whether the geometric patterns observed on MIMIC-IV generalize across datasets with a weaker hierarchical structure.

\begin{figure}[!h] 
  \centering \includegraphics[width=1\linewidth]{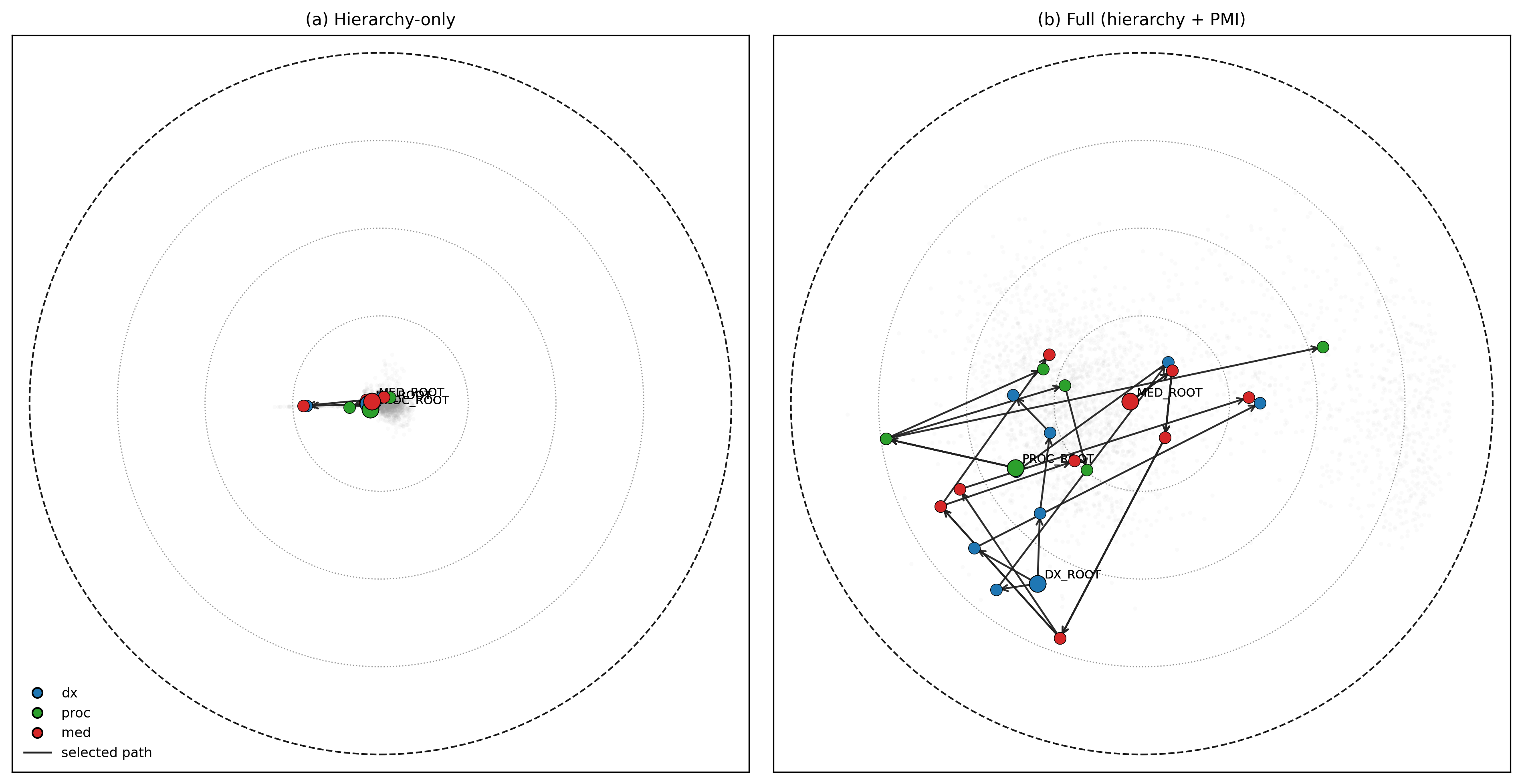} \caption{Hierarchy-only versus full hyperbolic embedding on eICU. Adding cross-modal PMI edges substantially improves geometric separation and local organization.} \label{fig:eicu_hierarchy_vs_full}
\end{figure}

Figure~\ref{fig:eicu_hierarchy_vs_full} compares embeddings learned using hierarchy edges alone versus the full graph with cross-modal PMI stitching. Similar to MIMIC-IV, the hierarchy-only embedding exhibits substantial geometric collapse, with most nodes concentrated near the origin and limited angular separation between branches. Adding PMI-based cross-modal edges produces a substantially more dispersed organization, introducing local clustering and improving separation across modalities and clinical contexts.

The effect is more pronounced on eICU than on MIMIC-IV, consistent with the weaker ontology structure of eICU, including synthetic procedure identifiers and less standardized medication coding. In this setting, hierarchy alone provides insufficient geometric constraints, while cross-modal temporal associations become critical for recovering meaningful latent structure.

\begin{figure}[!h] 
  \centering   \includegraphics[width=0.8\linewidth]{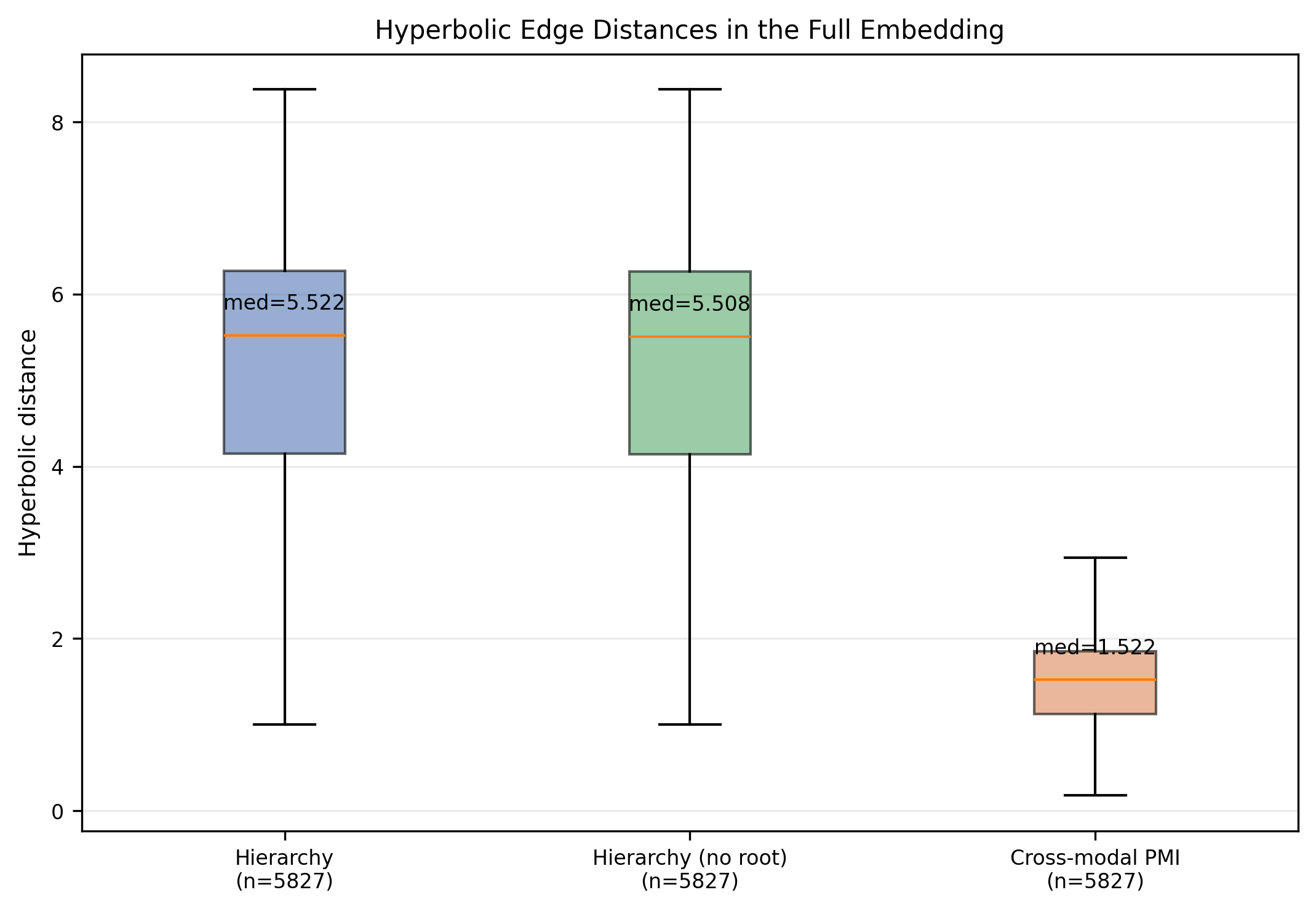} \caption{Hyperbolic distance distributions on eICU. PMI edges remain substantially more local than hierarchy edges in the learned space.} \label{fig:eicu_edge_dist}
\end{figure}

Figure~\ref{fig:eicu_edge_dist} further shows that hierarchy and PMI edges remain geometrically separated in the learned space. Hierarchy edges occupy substantially larger hyperbolic distances than PMI edges, indicating that the embedding preserves a distinction between global hierarchical organization and local associative structure even under noisier coding systems.

\section{Limitations and Ethical Considerations}
\label{app:limitations}

\paragraph{Limitations.}
Risk Horizons remains subject to several limitations inherent to longitudinal EHR modeling and structure-aware retrieval. First, clinical records are noisy, incomplete, and affected by coding variability across institutions and time, which can introduce inconsistencies in both hierarchical relations and temporal co-occurrence statistics. 

Second, the framework depends on retrieval quality and candidate-space coverage. If clinically relevant future events are absent from the retrieved Risk Horizon, downstream reranking cannot recover them, regardless of the strength of the inference model. This dependence is particularly important under sparse supervision or weak ontology structure, as observed in eICU.

Third, the approach depends on predefined coding systems and hierarchical organization. While hyperbolic geometry effectively captures structured relations, the quality of the learned embedding is influenced by the consistency and granularity of the underlying ontology. Datasets with synthetic or weakly standardized coding systems may therefore provide weaker geometric supervision.

Fourth, although LLM-based reranking improves predictive performance, it also introduces additional computational cost and potential variability across model backends. The proposed framework is intended as a research system for structured clinical reasoning rather than a deployable clinical decision-making tool, and its outputs should not be interpreted as medical recommendations without expert oversight and external validation.

Finally, Risk Horizons also inherits limitations of candidate-space retrieval: highly novel events or events arising from coding drift may fall outside the retrieved horizon. Addressing these cases may require adaptive retrieval expansion or dynamically updated retrieval structures under distributional shift.

\paragraph{Ethical Considerations.}
The use of EHR data raises important ethical concerns, including patient privacy, demographic bias, and potential misuse of predictive models. Although we use publicly available de-identified datasets, biases present in the underlying healthcare systems may still be reflected in the learned representations and predictions. Performance disparities may therefore emerge across demographic groups, institutions, or rare clinical conditions.

Risk Horizons aims to improve transparency and reduce unconstrained hallucination through structure-aware retrieval and constrained candidate generation. However, these mechanisms do not eliminate bias, guarantee fairness, or ensure clinical correctness. Any real-world deployment would require rigorous external validation, subgroup fairness analysis, prospective evaluation, and compliance with clinical and regulatory standards.



\end{document}